%% file: ms.tex
\pdfoutput=1

\documentclass[11pt]{article}

\usepackage[final]{acl}

\usepackage{times}
\usepackage{latexsym}

\usepackage[T1]{fontenc}

\usepackage[utf8]{inputenc}

\usepackage{microtype}

\usepackage{inconsolata}

\title{Guardians of the Machine Translation Meta-Evaluation:\\
Sentinel Metrics Fall In!}

\author{Stefano Perrella$^{1, *}$ \qquad Lorenzo Proietti$^{1, *}$ \qquad  Alessandro Scirè$^{1,2}$ \\ {\bf Edoardo Barba}$^1$ \qquad {\bf Roberto Navigli}$^1$ \\ $^1$Sapienza NLP Group, Sapienza University of Rome \\ $^2$Babelscape, Italy \\ \{perrella, lproietti, scire, barba, navigli\}@diag.uniroma1.it}

\usepackage{amsmath}
\usepackage{xspace}
\usepackage{tabularx}
\usepackage{graphicx}
\usepackage{booktabs}
\usepackage{xcolor}
\usepackage{color, colortbl}
\usepackage{hyperref}
\usepackage{subcaption}

\definecolor{lightgray}{gray}{0.9} 
\definecolor{lightblue}{rgb}{0.8,0.8,1.0}
\definecolor{lightred}{rgb}{1.0,0.8,0.8}
\definecolor{lightgreen}{rgb}{0.8,1.0,0.8}

\newcommand{\gemba}{GEMBA-MQM\xspace}
\newcommand{\matese}{MaTESe\xspace}
\newcommand{\comet}{COMET\xspace}
\newcommand{\xcomet}{XCOMET-Ensemble\xspace}
\newcommand{\xcometqe}{XCOMET-QE-Ensemble\xspace}
\newcommand{\metricx}{MetricX-23\xspace}
\newcommand{\metricxqe}{MetricX-23-QE\xspace}
\newcommand{\bleurt}{BLEURT-20\xspace}
\newcommand{\cometkiwi}{CometKiwi\xspace}
\newcommand{\candonly}{$\textsc{sentinel}_{\textsc{cand}}$\xspace}
\newcommand{\srconly}{$\textsc{sentinel}_{\textsc{src}}$\xspace}
\newcommand{\refonly}{$\textsc{sentinel}_{\textsc{ref}}$\xspace}
\newcommand{\sentmatese}{$\textsc{sentinel}_{\textsc{matese}}$\xspace}
\newcommand{\sentgemba}{$\textsc{sentinel}_{\textsc{gemba}}$\xspace}

\newcommand{\candonlyready}{$\textsc{sentinel}_{\textsc{cand}}$}
\newcommand{\srconlyready}{$\textsc{sentinel}_{\textsc{src}}$}

\newcommand{\langpair}[2]{$\textsc{#1}\rightarrow\textsc{#2}$}
\newcommand{\bs}[1]{\boldsymbol{#1}}

\newcommand{\acceq}{$\text{acc}_{\text{eq}}$\xspace}

\begin{document}
\maketitle

\def\thefootnote{*}\footnotetext{Equal contribution.}
\def\thefootnote{\arabic{footnote}}

\begin{abstract}
Annually, at the Conference of Machine Translation (WMT), the Metrics Shared Task organizers conduct the meta-evaluation of Machine Translation (MT) metrics, ranking them according to their correlation with human judgments. Their results guide researchers toward enhancing the next generation of metrics and MT systems.
With the recent introduction of neural metrics, the field has witnessed notable advancements. Nevertheless, the inherent opacity of these metrics has posed substantial challenges to the meta-evaluation process.
This work highlights two issues with the meta-evaluation framework currently employed in WMT, and assesses their impact on the metrics rankings. To do this, we introduce the concept of sentinel metrics, which are designed explicitly to scrutinize the meta-evaluation process's accuracy, robustness, and fairness. By employing sentinel metrics, we aim to validate our findings, and shed light on and monitor the potential biases or inconsistencies in the rankings. We discover that the present meta-evaluation framework favors two categories of metrics: i) those explicitly trained to mimic human quality assessments, and ii) continuous metrics. Finally, we raise concerns regarding the evaluation capabilities of state-of-the-art metrics, emphasizing that they might be basing their assessments on spurious correlations found in their training data.

\end{abstract}

\section{Introduction}
Over the past few years, the Machine Translation (MT) field has witnessed significant advancements, largely driven by the advent of neural architectures, with the Transformer \cite{NIPS2017_3f5ee243} being the most notable. Modern MT systems deliver mostly fluent and accurate translations, posing a challenge for their quality evaluation -- even when conducted by human annotators, especially those who lack professional training \cite{freitag-etal-2021-experts}. 
Under these circumstances, shallow overlap-based metrics are gradually being replaced by neural-based metrics, which demonstrate a better correlation with human judgments \cite{freitag-etal-2022-results}. 
However, a significant limitation is that most neural-based metrics are black-box systems trained to predict human judgments in the form of scalar scores, and typically do not provide justifications for their assessments. Besides rendering them challenging to interpret, such opacity also complicates their meta-evaluation. In this respect, we found that certain strategies for the assessment of MT metrics' capabilities -- which have recently been employed in the context of the Metrics Shared Task at the Conference on Machine Translation (WMT)\footnote{With its first edition in 2006 \cite{koehn-monz-2006-manual}, "WMT is the main event for machine translation and machine translation research." (\url{https://machinetranslate.org/wmt}).} -- favor specific metric categories and potentially encourage undesirable metrics behavior. To demonstrate these problems, we introduce the concept of sentinel metrics, i.e., a suite of metrics serving as a probe to identify pitfalls in the meta-evaluation process. Sentinel metrics are either trained with incomplete information -- which makes them inherently unable to evaluate the quality of machine-translated text properly -- or consist of variations of existing metrics -- which have been devised to expose specific issues in the meta-evaluation.

As an example, in Table~\ref{tab:official-ranking}, we present the segment-level ranking of WMT23 with the inclusion of a sentinel metric. As can be seen, \candonly ranks in the upper half. \candonly is a sentinel metric designed to assess the quality of a candidate translation based solely on the translation itself, without accessing its source sentence or any reference translation. Arguably, such a metric should only be capable of evaluating a translation's fluency, but not its adequacy in conveying the original message, and a fair assessment should rank it at lower positions. Notably, \candonly is above strong baselines such as \comet \cite{rei-etal-2020-comet} and \bleurt \cite{sellam-etal-2020-bleurt},  suggesting that there might be some issues with the segment-level meta-evaluation methods used in WMT23. 

\input{tables/official-ranking}

In this work, we: i) illustrate the issues that affect the segment-level meta-evaluation measures used in WMT23, demonstrating their impact experimentally with the help of sentinel metrics; ii) propose solutions for addressing these issues; iii) raise concerns regarding the reliability of state-of-the-art MT metrics. 
We publish the code to reproduce our work and the weights of the sentinel metrics at \url{https://github.com/SapienzaNLP/guardians-mt-eval}.

\section{The Meta-evaluation of MT Metrics} \label{sec:meta-evaluation}
Yearly, the WMT Metrics Shared Task organizes a competition among metrics, including participants' submissions and baselines, to identify the metric that most closely aligns with human judgments. Historically, the organizers have employed correlation with human judgment as a meta-evaluation strategy. Recently, significant efforts have been made to refine the meta-evaluation process, encompassing the adoption of new measures, such as those proposed by \citet{kocmi-etal-2021-ship} and \citet{deutsch-etal-2023-ties}, and the introduction of the challenge sets sub-task \cite{freitag-etal-2021-results, freitag-etal-2022-results}, among other initiatives. 
In this section, we provide an overview of WMT's official meta-evaluation setting.

First, multiple MT systems are employed to translate source segments found in one or more test datasets.\footnote{A segment typically refers to a single sentence, but can also include multiple sentences. For instance, at WMT23, a segment represents an entire paragraph rather than a single sentence for the English-to-German translation direction.} Consequently, test datasets contain several translations of the same source segment. Second, a manual evaluation campaign is carried out to assess the quality of all translations. Finally, metrics' capabilities are assessed based on their alignment with human judgments, which are in the form of scalar scores. Such alignment is typically estimated using correlation and accuracy measures. Specifically, metrics are evaluated at two granularity levels: 
\begin{itemize}
    \item at the segment level, metrics assign a score to every translation, and they are ranked according to their ability to discern between higher- and lower-quality translations;

    \item at the system level, metrics assign a score to each MT system,\footnote{Typically, the score of a system is calculated as the mean of the scores given to its translations.} and they are ranked according to their ability to discern between superior and inferior systems.
\end{itemize}
At both granularity levels, metrics can be evaluated using several statistical methods, such as the Kendall $\tau$ and Pearson $\rho$ correlation coefficients, which have traditionally been applied at the segment and system levels, respectively. A final metrics ranking is derived by aggregating results from all the chosen statistics. For example, at WMT23, the final ranking was computed from the following three statistics:
\begin{enumerate}
    \item System-level pairwise ranking accuracy \cite{kocmi-etal-2021-ship}, which evaluates metrics based on their ability to rank systems in the same order as human judgments.

    \item System- and segment-level Pearson correlation, which measures the degree to which metric scores and human scores are correlated linearly.

    \item Segment-level pairwise ranking accuracy with tie calibration \cite{deutsch-etal-2023-ties}, which evaluates metrics based on their ability to rank segments in the same order as human judgments, or their ability to predict ties correctly.
\end{enumerate}
In this work, we identify two critical issues related to the second and third statistics, and provide the following recommendations to address them:

\begin{itemize}
    \item \textbf{Translations should be grouped by their source segment before calculating segment-level correlations} (Section~\ref{sec:grouping}).

    \item \textbf{Tie calibration should not be conducted on the test set} (Section~\ref{sec:ties}). 
        
\end{itemize}
In the following two sections, we provide an overview of some of the aforementioned statistics, illustrate their flaws, and demonstrate their impact by leveraging our sentinel metrics.

\section{To Group or Not to Group?} \label{sec:grouping}
At early editions of the WMT Metrics Shared Task \cite{machacek-bojar-2013-results,machacek-bojar-2014-results,stanojevic-etal-2015-results,bojar-etal-2016-results}, human assessments were collected in the form of Relative Rankings (RR). Specifically, the annotators were tasked to rank up to $5$ translations of the same source sentence, produced by different MT systems. From each ranking, up to 10 pairwise comparisons were extracted. Despite metrics assessments being scalar scores -- which theoretically enabled the comparison of all pairs of translated segments -- correlation was measured only on those pairs of translations for which RR annotations were available. Therefore, only translations of the same source sentence were compared. 
Later on, at subsequent editions of WMT, new techniques for human evaluation were adopted: first, Direct Assessments \cite[DA]{graham-etal-2013-continuous} -- where annotators rate individual translations on a scale from 0 to 100 -- then,  Multidimensional Quality Metrics \cite[MQM]{ddd.uab.cat:130144} -- where annotators tag the spans of a translation that contain errors, specifying their category and severity. With both the new annotation schemas, each translated segment was assigned a scalar quality score independently of the other translations,\footnote{In MQM, a final score is obtained by applying a specific weighting to each combination of the detected spans' category and severity.} which made it possible to compare all translations, not only those of the same source sentence. This new possibility raised doubts regarding the best way to compute the correlation between metrics and human assessments. Indeed, it
could be computed using all translations at once -- \textit{No Grouping} -- or by first grouping translations based on either their source segment -- \textit{Segment Grouping} -- or the system that produced them -- \textit{System Grouping} -- and then returning the average correlation of these groups.

At the WMT21 Metrics Shared Task, \citet{freitag-etal-2021-results} chose the \textit{No Grouping} strategy, arguing that the other options would provide only a partial view of the overall picture. At WMT22, all three grouping strategies were used \cite{freitag-etal-2022-results}, and later at WMT23, \citet{freitag-etal-2023-results} chose \textit{No Grouping} again. Although \textit{No Grouping} is the only strategy that assesses the MT metrics' ability to discern between higher- and lower-quality translations in absolute terms, irrespective of the source segment or MT system, we show that both \textit{No Grouping} and \textit{System Grouping} may introduce unfairness and favor trained metrics over the rest. 

\subsection{The Relation Between Spurious Correlations and Grouping Strategies} \label{sec:problem-none-sys-grouping}
Most neural-based metrics are trained with a regression objective to approximate human judgments. They are expected to infer by pattern-matching the relation between human judgments and various phenomena, such as omissions, additions, or other translation errors. However, this mechanism might inadvertently lead to the detection of patterns that are not in a causal relation with the concept of translation quality, but are instead spurious correlations, e.g., the length of a translation, or the number of named entities in it, among others. Arguably, the meta-evaluation should not reward metrics for basing their assessments on spurious correlations between the features of the source, translation, or reference, and the human judgments. However, our intuition is that \textit{No Grouping} and \textit{System Grouping} strategies might be doing so by allowing the comparison of translations from different sources.
To simplify, consider a metric that unfairly penalizes a translation solely because it contains many named entities. Using \textit{No Grouping} or \textit{System Grouping}, such a metric might have a non-negative correlation with human judgments if, on average, translating sentences containing many named entities is more challenging than translating other sentences, because MT systems would be making more mistakes in translating them. Therefore, exploiting such a pattern might be beneficial even though it is not causally related to the quality of a translation. In contrast, when using \textit{Segment Grouping}, such a pattern would be ineffective, as different translations of the same source sentence should contain the same amount of named entities. More generally, we would expect \textit{Segment Grouping} to lessen the impact of most spurious correlations derived from features shared by a source sentence and its translations. 

To assess the extent of this issue, we incorporate three sentinel metrics into the current meta-evaluation framework and re-compute the metrics' rankings using all grouping strategies. Crucially, we find that the impact of spurious correlations when \textit{No Grouping} and \textit{System Grouping} strategies are employed is substantial -- favoring trained metrics over the rest\footnote{Indeed, overlap-based metrics such as BLEU \cite{papineni-etal-2002-bleu} and chrF \cite{popovic-2015-chrf}, or LLM-based metrics such as \gemba \cite{kocmi-federmann-2023-gemba}, were not trained to mimic human assessments and should not be able to leverage spurious correlations. } -- and is significantly reduced with \textit{Segment Grouping}.

\subsection{The Sentinel Metrics}
This section describes the three sentinel metrics employed to measure the impact of grouping strategies on the meta-evaluation process:
\begin{enumerate}
    \item \candonly, which assesses the quality of a translation without taking its source or reference as input.
    \item \srconly, which predicts the quality of a translation solely based on its source.
    \item \refonly, which predicts the quality of a translation solely based on its reference.
\end{enumerate}
Having no information regarding the translation to evaluate, \srconly and \refonly can only learn spurious correlations between the features of the source and reference sentences, respectively, and the human judgments. \candonly, instead, is a metric with partial information. Indeed, it is possible to evaluate a translation's fluency and grammatical correctness without comparing it with its source or reference sentences, but not its adequacy. Nonetheless, we expect \candonly to base its assessments on spurious correlations also.

\subsection{Experimental Setup}
Sentinel metrics employ XLM-RoBERTa large \cite{conneau2020unsupervised} as their backbone model, with a multi-layer fully-connected neural network on top of the [CLS] token, which is used to output predictions in the form of scalar scores. We train sentinel metrics to minimize the Mean Squared Error (MSE) between their predicted scores and human judgments. Our dataset comprises a selection of data from WMT spanning 2017 to 2022, incorporating Direct Assessments (DA) and Multidimensional Quality Metrics (MQM) scores. Following \citet{rei-etal-2022-comet}, we train sentinel metrics for a single epoch using DA from 2017 to 2020 and fine-tune them for a further epoch using MQM data. Additional details regarding the training process are reported in Appendix~\ref{apx:exp-setup}.

\subsection{Results} \label{sec:grouping-results}
In Table~\ref{tab:grouping-zhen}, we report the ranking derived from the segment-level Pearson correlation of the primary submissions to the Metrics Shared Task of WMT23, with the inclusion of sentinel metrics, in the language direction \langpair{zh}{en}, and with all three grouping strategies. We report in Appendix~\ref{apx:grouping} the rankings alongside the correlation values for all the official translation directions of the Metrics Shared Task, i.e., \langpair{zh}{en}, \langpair{en}{de} and \langpair{he}{en}. As can be seen, \srconly ranks fourth and third when the grouping strategies are \textit{No Grouping} and \textit{System Grouping}, respectively, surpassing strong baselines like \comet or \bleurt, and even state-of-the-art metrics like \gemba. The only metrics that are not surpassed are large regression-based systems such as \xcomet \cite{guerreiro2023xcomet} and \metricx \cite{juraska-etal-2023-metricx}, which might have learned the same spurious correlations leveraged by the sentinel metrics, in addition to non-spurious patterns (cf. Section \ref{sec:correlations}).
Conversely, when grouping by segment, \srconly and \refonly are correctly positioned at the bottom of the ranking,\footnote{This had to be expected, given that both these metrics return the same assessment for all translations of the same source segment.} and \candonly ranks $11$th, compared to $3$rd and $2$nd with \textit{No Grouping} and \textit{System Grouping}, respectively. A notable difference between the grouping strategies is the positioning of \gemba, which is ranked $7$th and $9$th with \textit{No Grouping} and \textit{System Grouping}, respectively, and becomes first with \textit{Segment Grouping}. We hypothesize that this is due to \gemba being based on GPT-4, which has not been explicitly fine-tuned on human assessments and is less likely to leverage spurious correlations such as those described in Section~\ref{sec:problem-none-sys-grouping}. Interestingly, with grouping strategies other than \textit{Segment Grouping}, \gemba is surpassed by all the sentinel metrics.

\input{tables/pearson-grouping-zhen}

\candonly is the only sentinel metric that does not rank at the very bottom with \textit{Segment Grouping}, outperforming prismSrc \cite{thompson-post-2020-automatic} and embed\_llama \cite{dreano-etal-2023-embed}, and positioning itself within the same cluster of statistical significance as BLEU. This suggests that focusing solely on the candidate translation -- specifically, its fluency and grammatical correctness -- may be sufficient to exceed the performance of some less effective metrics, at least in terms of Pearson correlation with human judgments. Furthermore, we highlight that our results may provide an answer to the open question left at WMT23 regarding the inconsistency of segment-level and system-level correlations for prismSrc. \citet{freitag-etal-2023-results} noticed that, despite displaying a moderate correlation at the segment level, prismSrc was showing negative correlation values at the system level. As can be seen from Table~\ref{tab:grouping-zhen}, prismSrc ranks $15$th out of $24$ with \textit{No Grouping} but $13$th out of $14$ with \textit{Segment Grouping} (i.e., it is in the second to last significance cluster, close to the sentinel metrics). This result is consistent with prismSrc's negative correlation at the system level. 

In Appendix~\ref{apx:grouping}, we also report the rankings and correlations obtained using the Kendall $\tau$ correlation coefficient for each grouping strategy, to show that our findings are independent of the correlation measure, at least among those typically employed at WMT, i.e., Pearson $\rho$ and Kendall $\tau$.

\subsubsection{Are MT metrics learning from spurious correlations?} \label{sec:correlations}
We hypothesize that some of the trained metrics may be basing their assessments on the same spurious correlations as those leveraged by the sentinel metrics. To delve deeper into this, we measure their segment-level Pearson correlation with the sentinel metrics using \textit{No Grouping}. Surprisingly, \xcomet, \xcometqe, \metricx, and \metricxqe, which are the only metrics that surpass the sentinels in Table~\ref{tab:grouping-zhen}, display a high correlation with all three sentinel metrics. Interestingly, their correlation with \srconly is $0.750$, $0.736$, $0.690$, and $0.712$ (Figure~\ref{fig:correlations-zhen}), respectively, while their correlation with human judgment is $0.650$, $0.647$, $0.625$, and $0.647$, respectively (Table~\ref{tab:full-grouping-zhen}).
We recognize that these metrics share many similarities with our sentinels, as both are neural transformer-based systems and both were trained with the same regression-based objective, using largely the same data.  This similarity likely contributes to the high correlation values observed. However, with access limited to only the source segment, \srconly relies exclusively on spurious correlations to conduct the evaluation. For this reason, we argue that these results raise concerns about the reliability of state-of-the-art MT metrics, which may be learning to exploit spurious correlations to minimize the Mean Squared Error with human judgments during training. To further support our hypothesis, we plot in Figure~\ref{fig:length-bias-zhen} the relation between the assessments of \xcomet and translation length, which serves as a simple spurious correlate of translation quality.\footnote{We expect that longer sentences are, on average, more challenging to translate. Therefore, we anticipate that MT metrics might have learned to assign lower scores to longer translations, despite the length and quality of translations not being causally related.} We also plot the distribution of MQM human judgments over translation length. As we can see from the figure, \xcomet scores decrease at increasing candidate lengths, with the metric almost never assigning scores higher than $0.9$ to translations longer than $400$ characters. However, the distribution of human judgments shows that human annotators rated many of those translations as perfect or near-perfect, indicating that \xcomet might be biased to assign lower scores to longer translations, irrespective of their quality. 
Furthermore, the least-squares regression lines show that, on average, and as expected, longer translations contain more errors than shorter ones, and therefore are assigned lower scores by human annotators. This suggests that detecting biases of this type might be particularly complex without datasets crafted specifically for it.  

We leave the investigation of these phenomena to future work and, for further details, we direct readers to Appendix~\ref{apx:correlations}, where we report the pairwise correlation between most of the considered metrics and sentinel metrics, and to appendix~\ref{apx:length-bias}, where we report the relation between such metrics' assessments and translation length.

\input{figures/length-bias/combined-zhen}

\section{The Evaluation of Ties} \label{sec:ties}
In this Section, we focus on the third statistic among those described in Section~\ref{sec:meta-evaluation}, i.e., the segment-level pairwise ranking accuracy with tie calibration, dubbed \acceq by \citet{deutsch-etal-2023-ties}. Prior to WMT23, the organizers of the Metrics Shared Task used to employ the Kendall $\tau$ coefficient -- which is a statistic used to estimate the rank-based agreement between two sets of measurements \cite{Kendall1945TheTO} -- to measure the correlation between metrics and human judgments at the segment level. \citet{deutsch-etal-2023-ties} pointed out that the Kendall $\tau$ coefficient does not account for metrics correctly predicting ties,\footnote{Given a pair of translations whose quality has been assessed by human annotators, the pair is tied if both translations were assigned with the same score.} and introduced \acceq to address this issue.
Unfortunately, our analysis indicates that \acceq inadvertently compromises evaluation fairness in order to accommodate ties, ultimately biasing the results in favor of continuous metrics\footnote{By continuous, we refer to those metrics whose assessments can take on any value within a given range, as opposed to discrete metrics, which can take on a limited set of values. Metrics from the COMET family such as COMET, XCOMET-Ensemble, and CometKiwi \cite{rei-etal-2022-cometkiwi} are continuous, whereas GEMBA-MQM \cite{kocmi-federmann-2023-gemba} and MaTESe \cite{perrella-etal-2022-matese} are examples of discrete metrics.} over discrete ones. 

\subsection{The Kendall \texorpdfstring{$\boldsymbol{\tau}$}{tau}}
In this section, we define the Kendall $\tau$ coefficient as employed by the organizers of the Metrics Shared Task of WMT21 and WMT22.\footnote{This is $\tau_b$ in \citet{deutsch-etal-2023-ties}.} Let $\bs{m}, \bs{h}$ be the vectors of metric and human assessments, respectively. \textit{Concordant} pairs are the pairs of metric assessments that have been ranked in the same order by humans; \textit{discordant} pairs are those ranked in a different order. We define $C$ and $D$ as the number of concordant and discordant pairs, respectively. We also define $T_h$ as the number of pairs only tied in the gold scores, $T_m$ as the number of pairs only tied in the metric scores, and $T_{hm}$ as the number of pairs tied both in gold and metric scores, i.e., the number of correctly predicted ties.
The Kendall $\tau$ correlation coefficient is defined as follows \cite{Kendall1945TheTO}:
\begin{equation}
    \tau = \frac{C - D}{\sqrt{(C + D + T_h)(C + D + T_m)}}.
    \label{eq:kendall}
\end{equation}

\subsection{The \texorpdfstring{\acceq}{acceq}}
As noted by \citet{deutsch-etal-2023-ties}, Kendall $\tau$ penalizes the prediction of ties, but never rewards them, as $T_m$ and $T_h$ are in the denominator, and $T_{hm}$ is not used. This issue was not prominent in the earliest editions of the Metrics Shared Task, where ties in human scores were disregarded, and older metrics rarely produced ties. Currently, instead, it is essential to consider the prediction of ties, especially since human MQM annotations contain a lot of them,\footnote{This is also due to the increasing quality of automatic translation, as perfect translations are assigned the same maximum score.} and some recently-proposed metrics are designed to output evaluation assessments that resemble MQM \cite{perrella-etal-2022-matese,kocmi-federmann-2023-gemba}. 
For this reason, \citet{deutsch-etal-2023-ties} proposed a measure that mimics the $\tau$ coefficient in the way it is computed, but also accounts for correctly predicting ties:
\begin{equation}
     \text{acc}_{\text{eq}} = \frac{C + T_{hm}}{C + D + T_h + T_m + T_{hm}}.
     \label{eq:acceq}
\end{equation}
Differently from Kendall $\tau$, \acceq includes $T_{hm}$ in the numerator, and the denominator encompasses the total number of pairs. Notably, discordant pairs are not subtracted from the numerator, rendering this metric a measure of accuracy, with scores ranging between $0$ and $1$. In Appendix~\ref{apx:kendall-example}, we provide a numerical example of the computation of both Kendall $\tau$ and \acceq from the vectors $\bs{m}$ and $\bs{h}$.

The \acceq measure, as it stands, would unfairly disadvantage continuous metrics. Indeed, it is extremely infrequent for such metrics to assign the same score to two different translations, meaning that they never predict ties. To address this issue, \citet{deutsch-etal-2023-ties} propose the tie calibration algorithm. In the following section, we briefly illustrate this algorithm and explain why it should not be conducted on the same test set used for the meta-evaluation.

\subsection{Tie Calibration} \label{sec:problem-tie-calibration}
The tie calibration algorithm determines, for each metric, a threshold $\epsilon$ such that, given two metric assessments $m_1$ and $m_2$, they are tied if $|m_1 - m_2| \le \epsilon$. \citet{deutsch-etal-2023-ties} propose selecting the $\epsilon$ that maximizes \acceq on the same test set used for the metrics meta-evaluation, enabling metrics to output the number of tied scores that best fits the distribution of human ties in the considered test set. This distribution is not stable across test sets (Table~\ref{tab:ties-testsets}), and \citet{deutsch-etal-2023-ties} show that $\epsilon$ values are not stable either. Nonetheless, they argue that this would not impact the fairness of the evaluation. 
Unfortunately, our analysis shows that this is not the case. Specifically, despite all metrics' $\epsilon$ values being selected on the same test data, we demonstrate that continuous metrics are more flexible to best fit the underlying distribution of human ties, compared to discrete ones, leading to unfairly higher \acceq values.

\input{figures/accuracy-and-threshold-varying-ties-grey}

\subsection{Two New Sentinel Metrics} \label{sec:new-sentinels}
To demonstrate the impact of this phenomenon, we introduce two additional sentinel metrics, i.e., \sentgemba and \sentmatese. \gemba \cite{kocmi-federmann-2023-gemba} and \matese \cite{perrella-etal-2022-matese} are MT metrics that output discrete scores in the form of MQM quality assessments and participated in WMT23. \sentgemba and \sentmatese are perturbed versions of \gemba and \matese, respectively, obtained by adding Gaussian noise -- $\mathcal{N}(0, 0.0001)$ -- to their predictions. By making their output continuous in the neighborhood of discrete values, we partially fill their gap with continuous metrics, while preventing any two different discrete assessments from inverting their ordering. That is, if two \gemba's assessments $m_1, m_2$ are such that $m_1$ > $m_2$, this relation is preserved by \sentgemba. In general, we expect a fair meta-evaluation to rank these sentinels on par or below their discrete counterparts.
Furthermore, we wish to remark that this solution is sub-optimal compared to metrics that are continuous by design. Indeed, due to the addition of Gaussian noise, the ordering of all \sentgemba and \sentmatese's assessments in the neighborhood of discrete values is randomized. 
 
To demonstrate that \sentgemba and \sentmatese can better fit the distribution of human ties compared to their discrete counterparts, we modify such a distribution in the test data.
Specifically, we repeatedly sub-sample the test data, such that for each pair of tied human assessments we remove that pair from the test data with a certain probability $p_t$, and do the same for non-tied pairs, which are removed with probability $p_n$.
We extract $13$ samples by assigning various values to $p_t$ and $p_n$ and report the chosen values in Table~\ref{tab:filtering-probs-zhen} in Appendix~\ref{apx:ties}.
As a consequence, each pair $(p_t, p_n)$ represents a different sub-sample of test data, with a different percentage of tied human pairs. Then, for each metric, we select the best $\epsilon$ and compute \acceq on each of these samples. 

\subsection{Results} \label{sec:results-ties}
In Figure~\ref{fig:accuracy-and-threshold-varying-ties-grey} (left), we present the \acceq results for a subset of continuous metrics, together with \gemba, \matese, \sentgemba, and \sentmatese. We discuss our results on the WMT23 \langpair{zh}{en} test set, and report results concerning the other language directions, i.e., \langpair{en}{de} and \langpair{he}{en}, in Appendix~\ref{apx:ties}. At first glance, it is evident that discrete metrics exhibit a distinct \acceq pattern compared to continuous and sentinel metrics. Notably, at lower percentages of tied human pairs, \sentgemba and \sentmatese significantly outperform \gemba and \matese.\footnote{It is important to highlight that the range of human tie percentages explored in our analysis is similar to that found in the WMT test sets. Indeed, as shown in Table~\ref{tab:ties-testsets}, such percentages range from a minimum of $15.14\%$ to a maximum of $53.35\%$, observed in the WMT22 \langpair{en}{de} test set.} This discrepancy arises because the tie calibration algorithm selects very small $\epsilon$ values, close to $0$ for every metric, allowing the number of ties predicted by continuous metrics to potentially drop to $0$. Conversely, metrics that yield discrete scores inherently produce a certain number of ties, placing them at a disadvantage, and thus ranking conceptually identical metrics like \sentgemba and \gemba at significantly different positions. Interestingly, in the hypothetical scenario in which there are no tied human pairs in the dataset, \sentgemba would rank second (despite several of its assessments having a random ordering), whereas \gemba would be second to last.  At increasing percentages of gold ties, instead, the \acceq values obtained by \sentgemba and \sentmatese converge to those of their discrete counterparts. However, this is a limitation of these sentinels' design and does not imply that the evaluation is fair at higher percentages of human ties. 

To better investigate the source of unfairness, in Figure~\ref{fig:accuracy-and-threshold-varying-ties-grey} (right) we show how the optimal $\epsilon$ changes at varying percentages of human ties. As can be seen, continuous metrics' $\epsilon$ is dynamically adjusted with heightened sensitivity, contrary to what happens for discrete metrics. Specifically, their $\epsilon$ is exactly $0$ until the percentage of human ties over all pairs is $39\%$. Additionally, for \matese, it remains constant between $44\%$ and $56\%$, and between $61\%$ and $68\%$, and the same happens for \gemba between $47\%$ and $51\%$ and between $56\%$ and $68\%$. In contrast, the values change for all the other metrics in the same intervals, enabling them to better fit the distribution of gold ties found in the test set.

\subsubsection{Can we use a held-out set for tie calibration?}
We have demonstrated that conducting the tie calibration on the same test set used for the evaluation favors continuous metrics over discrete ones. Nonetheless, this does not necessarily mean using a held-out dataset would ensure a fair meta-evaluation.
Indeed, our experiments show that unfairness stems from the different levels of adaptability between continuous and discrete metrics to the distribution of human ties found in the dataset used for tie calibration. Therefore, we expect that using a held-out dataset would still advantage continuous metrics if the distribution of human ties in the held-out resembled that of the test set, and disadvantage them if such a distribution differed from that of the test set. In both cases, continuous metrics' increased adaptability compared to discrete metrics would impair the fairness of the evaluation. To investigate this further, we compute a $80$-$20$ split of the test set to obtain an evaluation set for tie calibration. Then, we repeatedly sub-sample such an evaluation set to modify its distribution of human ties and compute \acceq on the new test set. The results are shown in Figure~\ref{fig:held-out-zhen}. We observe that the ranking is unstable at varying percentages of human ties, putting continuous metrics at a disadvantage if the proportion of ties in the evaluation set deviates significantly from that in the test set.

\input{figures/held-out-zhen}

\section{Conclusion}
In this work, we identified two issues with the current meta-evaluation of Machine Translation, as conducted at the Metrics Shared Task of the Conference on Machine Translation. We proposed a suite of sentinel metrics designed to highlight these issues and demonstrate their impact on the metrics rankings, revealing that certain metric categories are unfairly advantaged. Indeed, the \textit{None Grouping} and \textit{System Grouping} strategies favor trained metrics over overlap- and LLM-based ones and the algorithm of tie calibration favors continuous metrics over discrete ones, or vice versa, depending on the percentage of tied assessments in the dataset used for it. Specifically, continuous metrics are favored if the tie calibration is conducted on the same test set used for the evaluation.
Finally, we observed a notably high correlation between sentinel metrics and state-of-the-art metrics, raising concerns about their reliability and suggesting that their assessments might be based on spurious correlations present in the training data.

\clearpage

\section*{Acknowledgements}

\begin{center}
\noindent
    \begin{minipage}{0.1\linewidth}
        \begin{center}
            \includegraphics[scale=0.05]{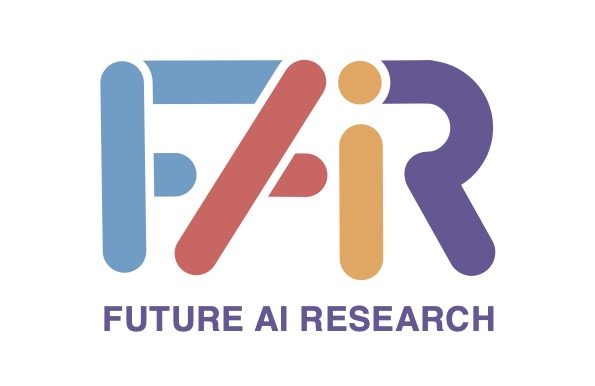}
        \end{center}
    \end{minipage}
    \hspace{0.01\linewidth}
    \begin{minipage}{0.70\linewidth}
         We gratefully acknowledge the support of the PNRR MUR project PE0000013-FAIR, and the CREATIVE project (CRoss-modal understanding and gEnerATIon of Visual and tExtual content), which is funded by the MUR Progetti di Rilevante Interesse Nazionale programme (PRIN 2020). 
    \end{minipage}
    \hspace{0.01\linewidth}
    \begin{minipage}{0.1\linewidth}
        \begin{center}
            \includegraphics[scale=0.08]{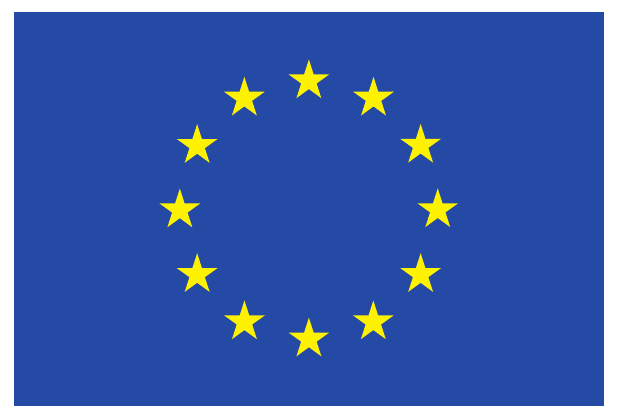}
        \end{center}
    \end{minipage}\\
\end{center}
\vspace{0.2cm}

\noindent This work has been carried out while Lorenzo Proietti and Alessandro Scirè were enrolled in the Italian National Doctorate on Artificial Intelligence run by Sapienza University of Rome.

\section*{Limitations}
Our analysis recommends grouping translations by their source segment before computing segment-level correlations with human judgments, showing that the rankings derived from the \textit{No Grouping} and \textit{System Grouping} strategies favor certain metric categories and potentially reward metrics for leveraging spurious correlations. However, we recognize that the \textit{Segment Grouping} strategy does not evaluate the ability of metrics to distinguish between higher- and lower-quality translations in absolute terms, that is, independently of their source sentence. We believe this aspect should play a role in the meta-evaluation process, and leave to future work the development of fairer methods to fill this gap. Furthermore, we acknowledge that, due to \textit{Segment Grouping}, each correlation measure is computed on a limited number of data points, i.e., as many as the MT systems that translated each source segment. In this respect, we argue that it would be necessary to investigate the metrics' ranking stability with varying numbers of MT systems, similar to the work of \citet{riley2024finding}, where they explored MT systems' ranking stability in designing human evaluation studies.  

Finally, we acknowledge that we did not provide a clear recommendation regarding a fair option for conducting the tie calibration algorithm. We demonstrated that continuous metrics are favored if selecting the optimal $\epsilon$ on the same test set used for the meta-evaluation and that using a held-out dataset would not be fair either. Nonetheless, using a held-out set would at least prevent the distribution of human ties used for tie calibration from being identical to that of the test set, and therefore it should be preferred. In general, we believe that a promising approach might involve studying the meaning of the score deltas of continuous metrics (akin to the work of \citet{kocmi2024navigating} regarding system-level assessments) and treating as tied all assessments within pre-defined score ranges derived from such deltas. This approach would also enhance the interpretability of MT metrics' assessments.

\bibliography{ms}

\appendix
\input{latex/appendix}

\end{document}

%% file: tables/official-ranking.tex
\begin{table}[!t]
    \centering
    \small
    \begin{tabular}{l r r}
    \toprule
    \textbf{Metric} &  & \multicolumn{1}{c}{\textbf{Avg. corr}} \\
    
    \cmidrule{2-3}
    
    XCOMET-Ensemble & $1$ & $0.697$ \\
    MetricX-23 & $2$ & $0.682$ \\
    XCOMET-QE-Ensemble* & $3$ & $0.681$ \\
    MetricX-23-QE* & $4$ & $0.681$ \\
    mbr-metricx-qe* & $5$ & $0.652$ \\
    GEMBA-MQM* & $6$ & $0.639$ \\
    MaTESe & $7$ & $0.636$ \\
    \underline{CometKiwi}* & $8$ & $0.632$ \\
    sescoreX & $9$ & $0.628$ \\
    \rowcolor{lightred}
    \candonlyready* & $10$ & $0.626$ \\
    cometoid22-wmt22* & $11$ & $0.625$ \\
    KG-BERTScore* & $12$ & $0.624$ \\
    \underline{COMET} & $13$ & $0.622$ \\
    \underline{BLEURT-20} & $14$ & $0.622$ \\
    Calibri-COMET22-QE* & $15$ & $0.603$ \\
    Calibri-COMET22 & $16$ & $0.603$ \\
    \underline{YiSi-1} & $17$ & $0.600$ \\
    \underline{docWMT22CometDA} & $18$ & $0.598$ \\
    \underline{docWMT22CometKiwiDA}* & $19$ & $0.598$ \\
    \underline{prismRef} & $20$ & $0.593$ \\
    \underline{MS-COMET-QE-22}* & $21$ & $0.588$ \\
    \underline{BERTscore} & $22$ & $0.582$ \\
    mre-score-labse-regular & $23$ & $0.558$ \\
    XLsim & $24$ & $0.544$ \\
    \underline{f200spBLEU} & $25$ & $0.540$ \\
    MEE4 & $26$ & $0.539$ \\
    tokengram\_F & $27$ & $0.537$ \\
    \underline{chrF} & $28$ & $0.537$ \\
    \underline{BLEU} & $29$ & $0.533$ \\
    \underline{prismSrc}* & $30$ & $0.530$ \\
    embed\_llama & $31$ & $0.529$ \\
    eBLEU & $32$ & $0.491$ \\
    \underline{Random-sysname}* & $33$ & $0.463$ \\
    \bottomrule
    \end{tabular}
    \caption{Segment-level ranking of the primary submissions to the WMT 2023 Metrics Shared Task, with the inclusion of sentinel metrics. The values in the column ``Avg. corr'' are obtained by averaging the correlations of the $6$ segment-level tasks of WMT 2023. Starred metrics are reference-free, underlined metrics are baselines, and highlighted metrics are sentinels. Ranks represent clusters of statistical significance and are computed following \citet{freitag-etal-2023-results}, which leverage the PERM-BOTH hypothesis test introduced by \citet{deutsch-etal-2021-statistical}. In Table~\ref{tab:full-official-ranking} in Appendix~\ref{apx:official-ranking}, we report the metrics' performance in terms of rank and correlation in all the $6$ tasks that contribute to this ranking. All the rankings present in this work have been computed using the official shared task library (\url{https://github.com/google-research/mt-metrics-eval}).}
    \label{tab:official-ranking}
\end{table}

%% file: tables/pearson-grouping-zhen.tex
\begin{table}[!ht]
\centering
\small
\begin{tabular}{lrrr}

\toprule

&   \multicolumn{3}{c}{\textbf{Grouping}} \\
\textbf{Metric} & \textbf{No} & \textbf{Seg} & \textbf{Sys} \\

\cmidrule{2-4}
XCOMET-Ensemble & $1$ & $2$ & $1$ \\
MetricX-23-QE* & $1$ & $4$ & $1$ \\
XCOMET-QE-Ensemble* & $1$ & $3$ & $1$ \\
MetricX-23 & $2$ & $3$ & $2$ \\
\rowcolor{lightred}
\candonlyready* & $3$ & $11$ & $2$ \\
\rowcolor{lightred}
\srconlyready* & $4$ & $14$ & $3$ \\
sescoreX & $4$ & $7$ & $5$ \\
MaTESe & $5$ & $6$ & $6$ \\
\rowcolor{lightred}
\refonly & $5$ & $14$ & $4$ \\
mbr-metricx-qe* & $6$ & $1$ & $7$ \\
cometoid22-wmt22* & $6$ & $4$ & $6$ \\
GEMBA-MQM* & $7$ & $1$ & $9$ \\
Calibri-COMET22-QE* & $7$ & $5$ & $8$ \\
\underline{CometKiwi}* & $7$ & $3$ & $9$ \\
KG-BERTScore* & $8$ & $4$ & $10$ \\
\underline{COMET} & $9$ & $4$ & $12$ \\
Calibri-COMET22 & $9$ & $7$ & $11$ \\
\underline{docWMT22CometKiwiDA}* & $10$ & $6$ & $13$ \\
\underline{BLEURT-20} & $10$ & $4$ & $13$ \\
\underline{MS-COMET-QE-22}* & $11$ & $7$ & $14$ \\
\underline{docWMT22CometDA} & $12$ & $6$ & $15$ \\
\underline{YiSi-1} & $13$ & $6$ & $16$ \\
\underline{BERTscore} & $14$ & $7$ & $17$ \\
\underline{prismSrc}* & $15$ & $13$ & $16$ \\
\underline{prismRef} & $16$ & $6$ & $18$ \\
embed\_llama & $17$ & $12$ & $18$ \\
mre-score-labse-regular & $18$ & $8$ & $19$ \\
\underline{BLEU} & $19$ & $11$ & $20$ \\
XLsim & $19$ & $10$ & $21$ \\
\underline{f200spBLEU} & $20$ & $10$ & $21$ \\
MEE4 & $20$ & $9$ & $21$ \\
\underline{chrF} & $21$ & $8$ & $22$ \\
tokengram\_F & $22$ & $8$ & $23$ \\
\underline{Random-sysname}* & $23$ & $14$ & $23$ \\
eBLEU & $24$ & $10$ & $24$ \\
\bottomrule
\end{tabular}
\caption{Rankings obtained from the segment-level Pearson correlation for the primary submissions to the WMT 2023 Metrics Shared Task, with sentinel metrics. The language direction is \langpair{zh}{en}. Ranks represent clusters of statistical significance. Additional information can be found in Appendix~\ref{apx:grouping}.}
\label{tab:grouping-zhen}
\end{table}

%% file: figures/length-bias/combined-zhen.tex
\begin{figure}[!ht]
    \centering
    \includegraphics[width=0.9\columnwidth]{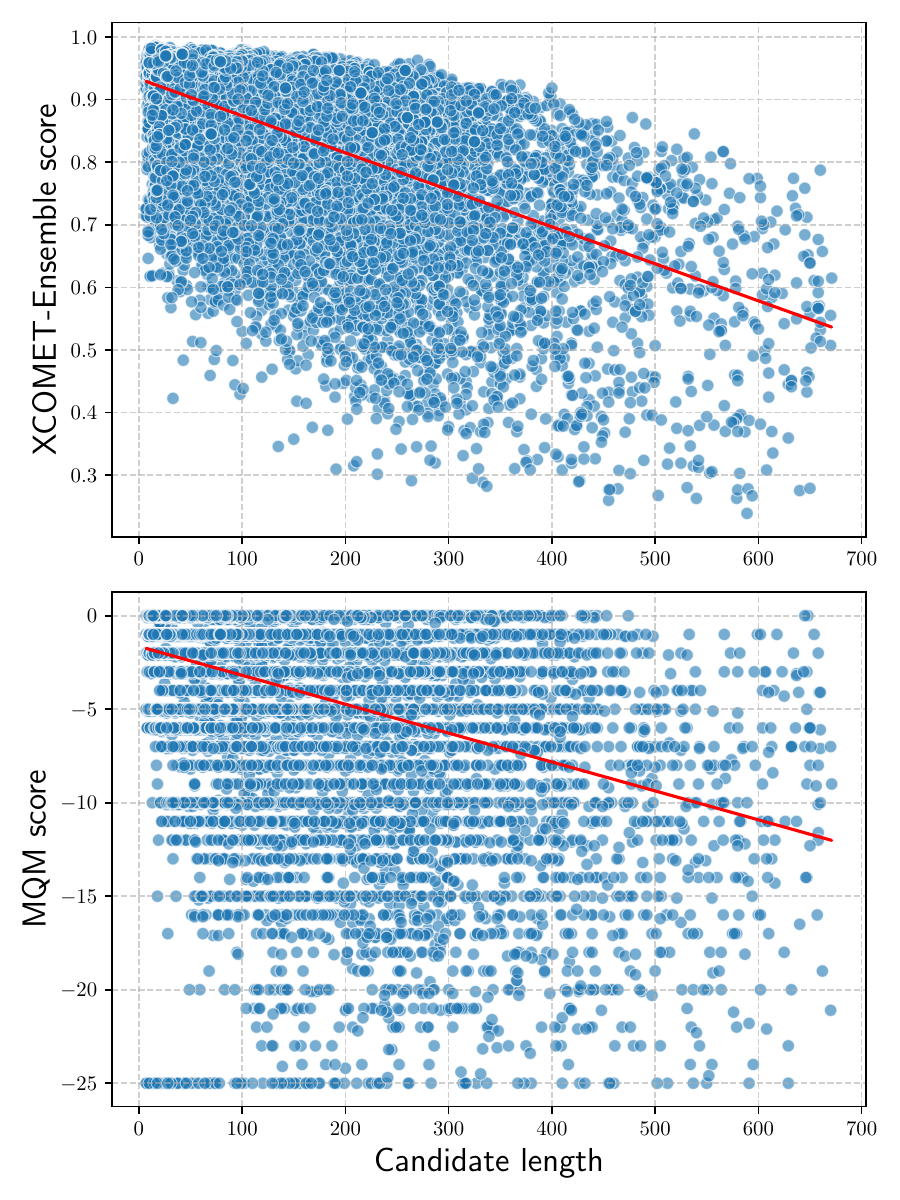}
    \caption{We show \xcomet assessments and MQM-based human judgments in the top and bottom figures, respectively, over the length of the candidate translation (in characters). The red line represents the linear least-squares regression. MQM human judgments smaller than $-25$ have been removed for improved clarity. The language pair is \langpair{zh}{en}.}
    \label{fig:length-bias-zhen}
\end{figure}

%% file: figures/accuracy-and-threshold-varying-ties-grey.tex
\begin{figure*}[!t]
    \centering
    \resizebox{\textwidth}{!}{
    \begin{subfigure}{\columnwidth}
        \includegraphics[width=0.9\columnwidth]{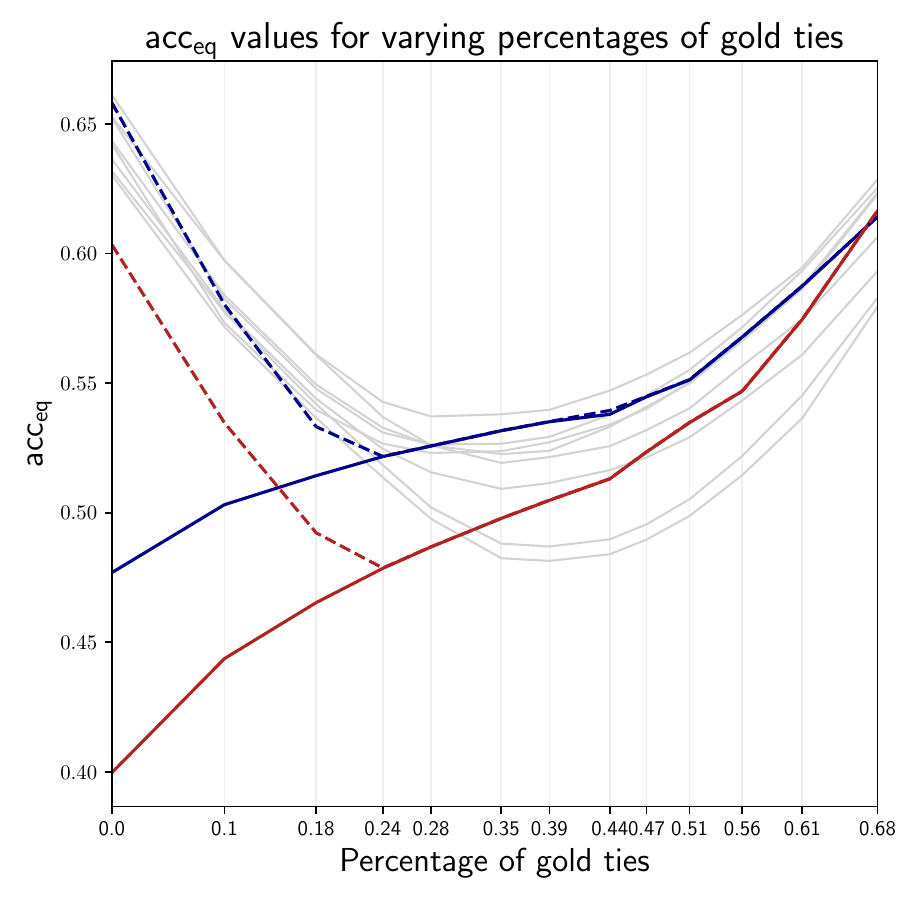}
        \label{fig:accuracy-varying-ties-grey}
    \end{subfigure}
    \begin{subfigure}{\columnwidth}
        \includegraphics[width=0.9\columnwidth]{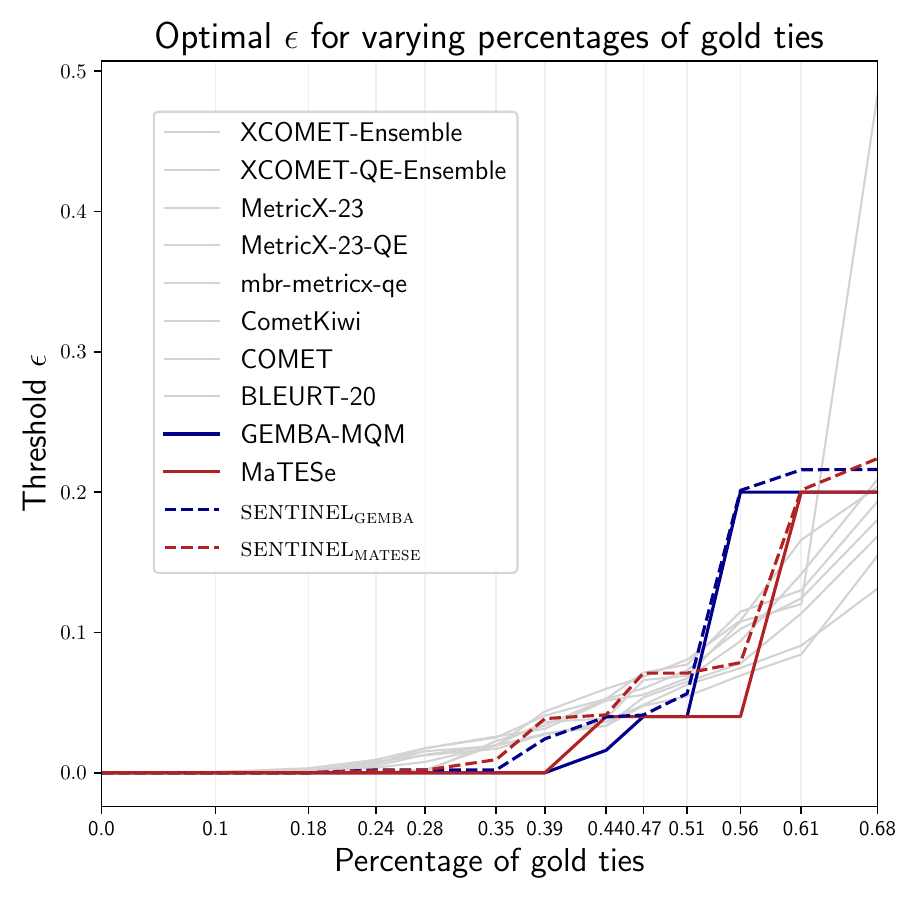}
        \label{fig:threshold-varying-ties-grey}
    \end{subfigure}
    }
    \caption{\acceq (left) and optimal $\epsilon$ (right) of the considered metrics for varying percentages of human ties in the test dataset, where $0.24$ is the percentage of human ties in the entire dataset, obtained when $p_t$ and $p_n$ are both $0$. $\epsilon$ values have been scaled using min-max scaling. Specifically, for each metric, the minimum $\epsilon$ is the optimal $\epsilon$ at $0\%$ of human ties, and the maximum is the optimal $\epsilon$ at $100\%$. The language direction is \langpair{zh}{en}. Results concerning all language directions can be found in Appendix~\ref{apx:ties}. For each percentage of human ties, we use $5$ different seeds to sub-sample the test data. Therefore, the shown \acceq and $\epsilon$, for each metric and percentage of ties, are averaged across $5$ different runs.}
    \label{fig:accuracy-and-threshold-varying-ties-grey}
\end{figure*}

%% file: figures/held-out-zhen.tex
\begin{figure}[!t]
    \centering
    \includegraphics[width=0.9\columnwidth]{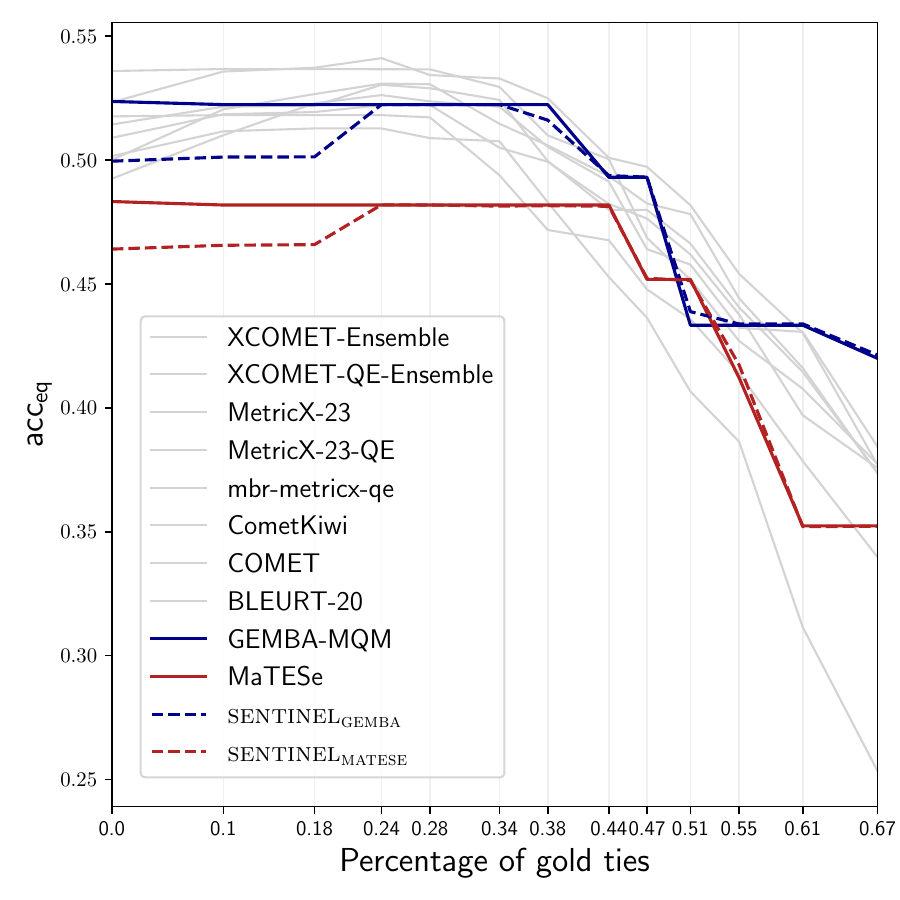}
    \caption{\acceq of the considered metrics when tie calibration is conducted on a held-out set, derived as a $20\%$ split of the test set, and repeatedly sub-sampled to modify its percentage of tied human scores. The x-axis represents the percentage of ties in the held-out set, while the y-axis represents the \acceq, as computed on the remaining $80\%$ of the test set. The language direction is \langpair{zh}{en}, and results concerning all language directions can be found in Appendix~\ref{apx:ties}. The percentage of human ties in the $80\%$ split of the test set is $24\%$. }
    \label{fig:held-out-zhen}
\end{figure}

%% file: latex/appendix.tex
\section{Official Ranking} \label{apx:official-ranking}
In Table~\ref{tab:full-official-ranking}, we report the official segment-level ranking of WMT23 Metrics Shared Task, including sentinel metrics.

\input{tables/full-official-ranking}

\section{Training the Sentinel Metrics} \label{apx:exp-setup}
The input for the sentinel metrics consists of either the source text (\srconly), candidate translation (\candonly), or reference translation (\refonly). Each sentence is tokenized and passed to the XLM-RoBERTa large model, which serves as a feature extractor. Then, we pass the embedding of the [CLS] token to a multi-layer, fully-connected neural network, which outputs the final scalar score.
More formally, considering $t$ as the input text for a sentinel metric:
\begin{align*}
\bs{e}_t &= \mathrm{XLM}\text{-}\mathrm{R}\left( t \right) \\
\bs{h}_t^{(1)} &= \mathrm{Dropout}\left( \mathrm{Tanh}\left( W_h^{(1)} \bs{e}_t + \bs{b}_h^{(1)} \right) \right) \\
\bs{h}_t^{(2)} &= \mathrm{Dropout}\left( \mathrm{Tanh}\left( W_h^{(2)} \bs{h}_t^{(1)} + \bs{b}_h^{(2)} \right) \right) \\
s_t &= W_o \bs{h}_t^{(2)} + \bs{b}_o
\end{align*}
Where:
\begin{itemize}
    \item $t$ is the tokenized input sentence.
    \item $\bs{e}_t$ is the [CLS] token embedding at the output of XLM-RoBERTa large.
    \item $\bs{h}_t^{(i)}$ represents the output of the $i^{th}$ layer of the fully-connected neural network. Each layer consists of a linear transformation, using weight matrix $W_h^{(i)}$ and bias vector $\bs{b}_h^{(i)}$, followed by a $\mathrm{Tanh}$ activation function and a dropout layer.
    \item $W_o$ and $\bs{b}_o$ are the output layer's weight matrix and bias vector, respectively.
    \item $s_t$ is the output scalar score assigned to sentence $t$.
\end{itemize}
Both training phases (i.e., the first, using DA-based human judgments, and the second, using MQM-based ones) employ the same set of hyperparameters, detailed in Table~\ref{tab:hyperparameters}.

\input{tables/hyperparameters}

\section{Grouping Strategies} \label{apx:grouping}
In Tables~\ref{tab:full-grouping-zhen}, \ref{tab:full-grouping-ende}, \ref{tab:full-grouping-heen}, we report the complete set of rankings and Pearson correlations, at the segment level, of the primary submissions to the WMT23 Metrics Shared Task, with sentinel metrics. Sentinel metrics are consistently ranked lower with \textit{Segment Grouping}.
Furthermore, in Tables~\ref{tab:full-kendall-grouping-zhen}, \ref{tab:full-kendall-grouping-ende}, \ref{tab:full-kendall-grouping-heen}, we report the complete set of rankings and Kendall $\tau$ correlation coefficients, at the segment level, of the primary submissions to the WMT23 Metrics Shared Task, with sentinel metrics. With Kendall $\tau$ as well, sentinel metrics rank lower when \textit{Segment Grouping} is employed.  We wish to note that \textit{Segment Grouping} requires the estimation of multiple correlation coefficients, which are then averaged. Consequently, each correlation is measured on a substantially smaller number of data points, compared to \textit{No Grouping} and \textit{System Grouping}. As a result, the number of clusters of statistical significance is reduced. Therefore, one should not focus on the absolute values of the ranks but on their value relative to that of the other metrics. For instance, in Table~\ref{tab:full-kendall-grouping-ende}, \candonly is ranked $5$th out of $19$ with \textit{No Grouping}, and $4$th out of $11$ with \textit{Segment Grouping}. While the absolute value of the rank is lower, in terms of correlation it has moved from the $8$th to the $17$th position. 

\input{tables/full-pearson-grouping-zhen}
\input{tables/full-pearson-grouping-ende}
\input{tables/full-pearson-grouping-heen}
\input{tables/full-kendall-grouping-zhen}
\input{tables/full-kendall-grouping-ende}
\input{tables/full-kendall-grouping-heen}

\section{Metrics Pairwise Correlations} \label{apx:correlations}
In Figures~\ref{fig:correlations-zhen}, \ref{fig:correlations-ende}, \ref{fig:correlations-heen}, we report the pairwise correlation between a subset of the primary submissions and baselines of WMT23, with the inclusion of sentinel metrics. We use Pearson correlation coefficient with \textit{No Grouping}. State-of-the-art regression-based metrics display a notably high correlation with sentinels. Specifically, the highest correlations are reported by \xcomet, \metricx, and their reference-less counterparts. Moderate correlation is also reported between sentinels and baseline metrics such as \cometkiwi, \comet, and \bleurt. As expected, instead, lexical-based metrics such as BLEU and chrF display close to no correlation with sentinels. Similarly, \gemba, a state-of-the-art LLM-based metric that has not been fine-tuned on human assessments, shows lower levels of correlation with the sentinel metrics, compared to the other state-of-the-art metrics.

\input{figures/correlations-zhen}
\input{figures/correlations-ende}
\input{figures/correlations-heen}

\section{Length Bias} \label{apx:length-bias}

In Figures~\ref{fig:length-bias-all-1},\ref{fig:length-bias-all-2} we report the relation between metrics assessments and the length of the candidate translation. We concatenate the data from all the three language directions used in the MQM-based evaluation of WMT23, i.e., \langpair{zh}{en}, \langpair{en}{de}, and \langpair{he}{en}. We wish to remind the reader that the meta-evaluation of WMT23 was conducted at the paragraph level for \langpair{en}{de}, and therefore, the reported candidate lengths are much larger than those in Figure~\ref{fig:length-bias-zhen}, which comprises only \langpair{zh}{en}. As we can see from the figures, most regression-based metrics, sentinels included, almost never assign very high scores to long translations, even if they are correct. This is in marked contrast to metrics trained with different objectives, such as \matese, or not fine-tuned to mimic the human judgment, such as \gemba. Indeed, both these metrics assign their highest score to several translations longer than $1200$ characters. Notably, there are several metrics whose assessments converge to a very narrow range of values as length increases. For example, \bleurt's assessments seem to be confined between approximately $0.4$ and $0.8$ for translations longer than $1000$ characters, and a similar pattern is observed for \comet.

\input{figures/length-bias/all-1}
\input{figures/length-bias/all-2}

\section{Kendall \texorpdfstring{$\boldsymbol{\tau}$}{tau} and \texorpdfstring{\acceq}{acceq} Computation Example} \label{apx:kendall-example}
In this section, we provide an example of the computation of Kendall $\tau$ and \acceq from two vectors of human and metric scores, i.e., $\bs{h}$ and $\bs{m}$ in the following table:
\vspace{5pt}
\begin{center}
    \begin{tabular}{lrrrr}
         \toprule
         $\bs{m}$ & 0.6 & 0.5 & 0.4 & 0.4 \\
         $\bs{h}$ & 5 & 3 & 5 & 5 \\
         \bottomrule
    \end{tabular}
\end{center}
\vspace{5pt}
For each vector, there are six pairs of assessments. In particular, the pairs of metric assessments are $(m_1,m_2)$, $(m_1,m_3)$, $(m_1,m_4)$, $(m_2,m_3)$, $(m_2,m_4)$, $(m_3,m_4)$. 

In Equations \ref{eq:kendall} and \ref{eq:acceq}, $C=1$, since the only concordant pair is $(m_1, m_2)$. Indeed, $m_1 > m_2$ and $h_1 > h_2$. $D=2$, since the pairs $(m_2,m_3), (m_2,m_4)$ are discordant. $T_m = 0$, since there are no pairs tied only in the metric scores. $T_h = 2$, since the pairs $(h_1,h_3), (h_1,h_4)$ are tied only in the human scores. $T_{hm} = 1$, since the remaining pair, i.e., $(m_3,m_4)$, is tied in both human and metric scores. In this example, $\tau = -0.258$ and $\text{acc}_{\text{eq}} = 0.333$.

\section{Ties} \label{apx:ties}
In Table~\ref{tab:ties-testsets}, we report the percentage of tied human pairs in the datasets used in recent editions of WMT.

\input{tables/ties-testsets}

In Tables~\ref{tab:filtering-probs-zhen}, \ref{tab:filtering-probs-ende}, \ref{tab:filtering-probs-heen}, we report the values of $p_t$ and $p_n$ used to sub-sample the \langpair{zh}{en}, \langpair{en}{de}, and \langpair{he}{en} test sets, respectively, to conduct the experiment described in Section~\ref{sec:new-sentinels}. We also report the corresponding percentage of human ties and total number of pairs, for each sample.

\input{tables/filtering-probs-zhen}
\input{tables/filtering-probs-ende}
\input{tables/filtering-probs-heen}

In Figures~\ref{fig:accuracy-and-threshold-varying-ties-zh-en}, \ref{fig:accuracy-and-threshold-varying-ties-en-de}, \ref{fig:accuracy-and-threshold-varying-ties-he-en}, we report the \acceq and optimal $\epsilon$ for each considered metric, in all three language directions considered at WMT 2023.

\input{figures/accuracy-and-threshold-varying-ties-zhen}
\input{figures/accuracy-and-threshold-varying-ties-ende}
\input{figures/accuracy-and-threshold-varying-ties-heen}

In Figure~\ref{fig:dev-accuracy-varying-ties}, we report the \acceq values of the considered metrics, as computed on a $80\%$ split of the test set. $\epsilon$ values have been estimated using a held-out set derived as a $20\%$ split of the entire test set. The held-out set is repeatedly sub-sampled to vary its percentage of tied human scores. Different percentage values are reported on the x-axis.

\input{figures/dev-accuracy-varying-ties}

%% file: tables/full-official-ranking.tex
\begin{table*}[t]
\centering
\resizebox{\textwidth}{!}{
\begin{tabular}{lrr|rr|rr|rr|rr|rr|rr}
\toprule
& \multicolumn{2}{c}{} & \multicolumn{4}{c}{\textbf{\langpair{en}{de}}} & \multicolumn{4}{c}{\textbf{\langpair{he}{en}}} & \multicolumn{4}{c}{\textbf{\langpair{zh}{en}}} \\
\textbf{Metric} & \multicolumn{2}{c}{\textbf{Avg. corr}} & \multicolumn{2}{c}{\textbf{Pearson}} & \multicolumn{2}{c}{\textbf{\acceq}} & \multicolumn{2}{c}{\textbf{Pearson}} & \multicolumn{2}{c}{\textbf{\acceq}} & \multicolumn{2}{c}{\textbf{Pearson}} & \multicolumn{2}{c}{\textbf{\acceq}} \\
\cmidrule{2-15}
XCOMET-Ensemble & $1$ & $0.697$ & $1$ & $0.695$ & $1$ & $0.604$ & $1$ & $0.556$ & $1$ & $0.586$ & $1$ & $0.650$ & $1$ & $0.543$ \\
MetricX-23 & $2$ & $0.682$ & $4$ & $0.585$ & $1$ & $0.603$ & $1$ & $0.548$ & $2$ & $0.577$ & $2$ & $0.625$ & $3$ & $0.531$ \\
XCOMET-QE-Ensemble* & $3$ & $0.681$ & $2$ & $0.679$ & $3$ & $0.588$ & $3$ & $0.498$ & $4$ & $0.554$ & $1$ & $0.647$ & $3$ & $0.533$ \\
MetricX-23-QE* & $4$ & $0.681$ & $3$ & $0.626$ & $2$ & $0.596$ & $2$ & $0.520$ & $3$ & $0.564$ & $1$ & $0.647$ & $4$ & $0.527$ \\
mbr-metricx-qe* & $5$ & $0.652$ & $4$ & $0.571$ & $3$ & $0.584$ & $5$ & $0.411$ & $4$ & $0.553$ & $6$ & $0.489$ & $2$ & $0.537$ \\
GEMBA-MQM* & $6$ & $0.639$ & $6$ & $0.502$ & $5$ & $0.572$ & $5$ & $0.401$ & $3$ & $0.564$ & $7$ & $0.449$ & $5$ & $0.522$ \\
MaTESe & $7$ & $0.636$ & $5$ & $0.554$ & $9$ & $0.528$ & $4$ & $0.459$ & $5$ & $0.550$ & $5$ & $0.511$ & $12$ & $0.479$ \\
CometKiwi* & $8$ & $0.632$ & $7$ & $0.475$ & $5$ & $0.569$ & $7$ & $0.387$ & $6$ & $0.544$ & $7$ & $0.442$ & $4$ & $0.525$ \\
sescoreX & $9$ & $0.628$ & $6$ & $0.519$ & $6$ & $0.563$ & $7$ & $0.385$ & $16$ & $0.484$ & $4$ & $0.536$ & $9$ & $0.499$ \\
\rowcolor{lightred}
\candonlyready* & $10$ & $0.626$ & $5$ & $0.561$ & $6$ & $0.562$ & $10$ & $0.339$ & $16$ & $0.483$ & $3$ & $0.580$ & $14$ & $0.473$ \\
cometoid22-wmt22* & $11$ & $0.625$ & $8$ & $0.441$ & $4$ & $0.578$ & $9$ & $0.365$ & $12$ & $0.515$ & $6$ & $0.479$ & $7$ & $0.515$ \\
KG-BERTScore* & $12$ & $0.624$ & $8$ & $0.451$ & $7$ & $0.556$ & $8$ & $0.382$ & $7$ & $0.537$ & $8$ & $0.430$ & $6$ & $0.516$ \\
COMET & $13$ & $0.622$ & $9$ & $0.432$ & $4$ & $0.574$ & $5$ & $0.401$ & $8$ & $0.532$ & $9$ & $0.396$ & $7$ & $0.514$ \\
BLEURT-20 & $14$ & $0.622$ & $7$ & $0.484$ & $5$ & $0.572$ & $8$ & $0.382$ & $11$ & $0.519$ & $10$ & $0.378$ & $6$ & $0.518$ \\
Calibri-COMET22-QE* & $15$ & $0.603$ & $9$ & $0.441$ & $12$ & $0.483$ & $6$ & $0.395$ & $13$ & $0.506$ & $7$ & $0.443$ & $10$ & $0.491$ \\
Calibri-COMET22 & $16$ & $0.603$ & $10$ & $0.413$ & $10$ & $0.522$ & $5$ & $0.401$ & $12$ & $0.515$ & $9$ & $0.396$ & $14$ & $0.474$ \\
YiSi-1 & $17$ & $0.600$ & $12$ & $0.366$ & $8$ & $0.542$ & $6$ & $0.395$ & $8$ & $0.529$ & $12$ & $0.290$ & $8$ & $0.504$ \\
docWMT22CometDA & $18$ & $0.598$ & $11$ & $0.394$ & $7$ & $0.559$ & $10$ & $0.339$ & $14$ & $0.497$ & $11$ & $0.353$ & $10$ & $0.493$ \\
docWMT22CometKiwiDA* & $19$ & $0.598$ & $8$ & $0.444$ & $8$ & $0.547$ & $12$ & $0.286$ & $15$ & $0.489$ & $9$ & $0.387$ & $10$ & $0.493$ \\
prismRef & $20$ & $0.593$ & $6$ & $0.516$ & $10$ & $0.518$ & $11$ & $0.319$ & $9$ & $0.528$ & $14$ & $0.183$ & $8$ & $0.504$ \\
MS-COMET-QE-22* & $21$ & $0.588$ & $13$ & $0.310$ & $8$ & $0.546$ & $12$ & $0.295$ & $14$ & $0.498$ & $10$ & $0.367$ & $9$ & $0.498$ \\
BERTscore & $22$ & $0.582$ & $13$ & $0.325$ & $9$ & $0.528$ & $10$ & $0.335$ & $12$ & $0.515$ & $13$ & $0.236$ & $9$ & $0.499$ \\
mre-score-labse-regular & $23$ & $0.558$ & $18$ & $0.111$ & $9$ & $0.530$ & $8$ & $0.378$ & $10$ & $0.522$ & $16$ & $0.145$ & $12$ & $0.481$ \\
XLsim & $24$ & $0.544$ & $14$ & $0.239$ & $9$ & $0.527$ & $14$ & $0.233$ & $17$ & $0.480$ & $17$ & $0.111$ & $15$ & $0.464$ \\
f200spBLEU & $25$ & $0.540$ & $14$ & $0.237$ & $9$ & $0.526$ & $14$ & $0.230$ & $19$ & $0.447$ & $18$ & $0.108$ & $13$ & $0.476$ \\
MEE4 & $26$ & $0.539$ & $17$ & $0.202$ & $9$ & $0.529$ & $13$ & $0.256$ & $20$ & $0.441$ & $18$ & $0.105$ & $12$ & $0.480$ \\
tokengram\_F & $27$ & $0.537$ & $16$ & $0.227$ & $10$ & $0.520$ & $14$ & $0.226$ & $18$ & $0.461$ & $20$ & $0.060$ & $11$ & $0.485$ \\
chrF & $28$ & $0.537$ & $15$ & $0.232$ & $10$ & $0.519$ & $15$ & $0.221$ & $18$ & $0.460$ & $19$ & $0.063$ & $11$ & $0.485$ \\
BLEU & $29$ & $0.533$ & $17$ & $0.192$ & $10$ & $0.520$ & $15$ & $0.220$ & $20$ & $0.442$ & $17$ & $0.119$ & $14$ & $0.472$ \\
prismSrc* & $30$ & $0.530$ & $9$ & $0.425$ & $13$ & $0.426$ & $16$ & $0.140$ & $20$ & $0.441$ & $13$ & $0.223$ & $17$ & $0.421$ \\
embed\_llama & $31$ & $0.529$ & $14$ & $0.250$ & $12$ & $0.483$ & $15$ & $0.215$ & $21$ & $0.430$ & $15$ & $0.161$ & $16$ & $0.447$ \\
\rowcolor{lightred}
\srconlyready* & $32$ & $0.512$ & $7$ & $0.469$ & $15$ & $0.231$ & $10$ & $0.334$ & $21$ & $0.428$ & $4$ & $0.540$ & $19$ & $0.240$ \\
\rowcolor{lightred}
\refonly & $33$ & $0.506$ & $8$ & $0.464$ & $15$ & $0.231$ & $11$ & $0.301$ & $21$ & $0.428$ & $5$ & $0.506$ & $19$ & $0.240$ \\
eBLEU & $34$ & $0.491$ & $20$ & $-0.011$ & $11$ & $0.512$ & $16$ & $0.131$ & $19$ & $0.445$ & $22$ & $-0.084$ & $14$ & $0.473$ \\
Random-sysname* & $35$ & $0.463$ & $19$ & $0.064$ & $14$ & $0.409$ & $17$ & $0.041$ & $21$ & $0.428$ & $21$ & $0.018$ & $18$ & $0.381$ \\
\bottomrule
\end{tabular}
}
\caption{Complete segment-level results for the primary submissions to the WMT 2023 Metrics Shared Task, with sentinel metrics.}
\label{tab:full-official-ranking}
\end{table*}

%% file: tables/hyperparameters.tex
\begin{table}[t]
\centering
\resizebox{\columnwidth}{!}{
    \begin{tabular}{lc}
    \toprule
    \textbf{Hyperparameter} & \textbf{Value} \\
    \midrule
    Optimizer & RAdam \cite{liu2021variance} \\
    Learning Rate & 1e-6 \\
    Number of Epochs & 1 \\
    Batch Size & 8 \\
    Accumulation Steps & 2 \\
    Dropout & 0.1 \\
    Dimension of $\bs{h}_t^{(1)}$ & 512 \\
    Dimension of $\bs{h}_t^{(2)}$ & 128 \\
    \bottomrule
    \end{tabular}
}

\caption{Hyperparameters used for both training phases of the sentinel metrics.}
\label{tab:hyperparameters}
\end{table}

%% file: tables/full-pearson-grouping-zhen.tex
\begin{table*}[ht]
\small
\centering
\begin{tabular}{lrr|rr|rr}

\toprule

\textbf{Metric} & \multicolumn{2}{c}{\textbf{No}} & \multicolumn{2}{c}{\textbf{Segment}} & \multicolumn{2}{c}{\textbf{System}} \\

\cmidrule{2-7}
XCOMET-Ensemble & $1$ & $0.650$ & $2$ & $0.421$ & $1$ & $0.610$ \\
MetricX-23-QE* & $1$ & $0.647$ & $4$ & $0.359$ & $1$ & $0.610$ \\
XCOMET-QE-Ensemble* & $1$ & $0.647$ & $3$ & $0.380$ & $1$ & $0.612$ \\
MetricX-23 & $2$ & $0.625$ & $3$ & $0.373$ & $2$ & $0.580$ \\
\rowcolor{lightred}
\candonlyready* & $3$ & $0.580$ & $11$ & $0.201$ & $2$ & $0.578$ \\
\rowcolor{lightred}
\srconlyready* & $4$ & $0.540$ & $14$ & $0.000$ & $3$ & $0.561$ \\
sescoreX & $4$ & $0.536$ & $7$ & $0.295$ & $5$ & $0.505$ \\
MaTESe & $5$ & $0.511$ & $6$ & $0.325$ & $6$ & $0.441$ \\
\rowcolor{lightred}
\refonly & $5$ & $0.506$ & $14$ & $0.000$ & $4$ & $0.525$ \\
mbr-metricx-qe* & $6$ & $0.489$ & $1$ & $0.436$ & $7$ & $0.431$ \\
cometoid22-wmt22* & $6$ & $0.479$ & $4$ & $0.357$ & $6$ & $0.446$ \\
GEMBA-MQM* & $7$ & $0.449$ & $1$ & $0.434$ & $9$ & $0.378$ \\
Calibri-COMET22-QE* & $7$ & $0.443$ & $5$ & $0.355$ & $8$ & $0.411$ \\
\underline{CometKiwi}* & $7$ & $0.442$ & $3$ & $0.388$ & $9$ & $0.388$ \\
KG-BERTScore* & $8$ & $0.430$ & $4$ & $0.369$ & $10$ & $0.374$ \\
\underline{COMET} & $9$ & $0.396$ & $4$ & $0.364$ & $12$ & $0.345$ \\
Calibri-COMET22 & $9$ & $0.396$ & $7$ & $0.311$ & $11$ & $0.360$ \\
\underline{docWMT22CometKiwiDA}* & $10$ & $0.387$ & $6$ & $0.340$ & $13$ & $0.320$ \\
\underline{BLEURT-20} & $10$ & $0.378$ & $4$ & $0.371$ & $13$ & $0.330$ \\
\underline{MS-COMET-QE-22}* & $11$ & $0.367$ & $7$ & $0.306$ & $14$ & $0.313$ \\
\underline{docWMT22CometDA} & $12$ & $0.353$ & $6$ & $0.327$ & $15$ & $0.291$ \\
\underline{YiSi-1} & $13$ & $0.290$ & $6$ & $0.329$ & $16$ & $0.237$ \\
\underline{BERTscore} & $14$ & $0.236$ & $7$ & $0.309$ & $17$ & $0.186$ \\
\underline{prismSrc}* & $15$ & $0.223$ & $13$ & $0.078$ & $16$ & $0.243$ \\
\underline{prismRef} & $16$ & $0.183$ & $6$ & $0.332$ & $18$ & $0.135$ \\
embed\_llama & $17$ & $0.161$ & $12$ & $0.138$ & $18$ & $0.139$ \\
mre-score-labse-regular & $18$ & $0.145$ & $8$ & $0.251$ & $19$ & $0.123$ \\
\underline{BLEU} & $19$ & $0.119$ & $11$ & $0.208$ & $20$ & $0.093$ \\
XLsim & $19$ & $0.111$ & $10$ & $0.218$ & $21$ & $0.069$ \\
\underline{f200spBLEU} & $20$ & $0.108$ & $10$ & $0.220$ & $21$ & $0.077$ \\
MEE4 & $20$ & $0.105$ & $9$ & $0.236$ & $21$ & $0.070$ \\
\underline{chrF} & $21$ & $0.063$ & $8$ & $0.263$ & $22$ & $0.020$ \\
tokengram\_F & $22$ & $0.060$ & $8$ & $0.262$ & $23$ & $0.015$ \\
\underline{Random-sysname}* & $23$ & $0.018$ & $14$ & $0.019$ & $23$ & $0.002$ \\
eBLEU & $24$ & $-0.084$ & $10$ & $0.219$ & $24$ & $-0.115$ \\
\bottomrule
\end{tabular}

\caption{Segment-level Pearson correlation for the primary submissions to the WMT23 Metrics Shared Task, with sentinel metrics. The language direction is \langpair{zh}{en}. Starred metrics are reference-free, underlined metrics are baselines, and highlighted metrics are sentinels. Ranks represent clusters of statistical significance and are computed following \citet{freitag-etal-2023-results}, which leverage the PERM-BOTH hypothesis test introduced by \citet{deutsch-etal-2021-statistical}.}
\label{tab:full-grouping-zhen}
\end{table*}

%% file: tables/full-pearson-grouping-ende.tex
\begin{table*}[ht]
\centering
\small
\begin{tabular}{lrrrrrr}
\toprule
\multicolumn{1}{l}{\textbf{Metric}} & \multicolumn{2}{c}{\textbf{No}} & \multicolumn{2}{c}{\textbf{Segment}} & \multicolumn{2}{c}{\textbf{System}} \\

\cmidrule{2-7}
XCOMET-Ensemble & $1$ & $0.695$ & $1$ & $0.538$ & $1$ & $0.676$ \\
XCOMET-QE-Ensemble* & $2$ & $0.679$ & $2$ & $0.507$ & $2$ & $0.658$ \\
MetricX-23-QE* & $3$ & $0.626$ & $2$ & $0.511$ & $3$ & $0.564$ \\
MetricX-23 & $4$ & $0.585$ & $2$ & $0.507$ & $4$ & $0.547$ \\
mbr-metricx-qe* & $4$ & $0.571$ & $1$ & $0.543$ & $3$ & $0.551$ \\
\rowcolor{lightred}
\candonlyready* & $5$ & $0.561$ & $6$ & $0.396$ & $5$ & $0.522$ \\
MaTESe & $5$ & $0.554$ & $8$ & $0.330$ & $4$ & $0.526$ \\
sescoreX & $6$ & $0.519$ & $3$ & $0.459$ & $6$ & $0.502$ \\
\underline{prismRef} & $6$ & $0.516$ & $7$ & $0.349$ & $4$ & $0.528$ \\
GEMBA-MQM* & $6$ & $0.502$ & $3$ & $0.482$ & $7$ & $0.446$ \\
\underline{BLEURT-20} & $7$ & $0.484$ & $2$ & $0.492$ & $7$ & $0.455$ \\
\underline{CometKiwi}* & $7$ & $0.475$ & $3$ & $0.463$ & $7$ & $0.451$ \\
\rowcolor{lightred}
\srconlyready* & $8$ & $0.469$ & $12$ & $0.000$ & $6$ & $0.502$ \\
\rowcolor{lightred}
\refonly & $8$ & $0.464$ & $12$ & $0.000$ & $6$ & $0.492$ \\
KG-BERTScore* & $8$ & $0.451$ & $4$ & $0.456$ & $8$ & $0.421$ \\
\underline{docWMT22CometKiwiDA}* & $9$ & $0.444$ & $5$ & $0.426$ & $9$ & $0.404$ \\
cometoid22-wmt22* & $9$ & $0.441$ & $2$ & $0.499$ & $9$ & $0.385$ \\
Calibri-COMET22-QE* & $9$ & $0.441$ & $5$ & $0.432$ & $8$ & $0.414$ \\
\underline{COMET} & $9$ & $0.432$ & $2$ & $0.508$ & $10$ & $0.363$ \\
\underline{prismSrc}* & $9$ & $0.425$ & $11$ & $0.102$ & $6$ & $0.487$ \\
Calibri-COMET22 & $10$ & $0.413$ & $3$ & $0.477$ & $10$ & $0.370$ \\
\underline{docWMT22CometDA} & $11$ & $0.394$ & $3$ & $0.484$ & $11$ & $0.310$ \\
\underline{YiSi-1} & $12$ & $0.366$ & $5$ & $0.404$ & $12$ & $0.284$ \\
\underline{BERTscore} & $13$ & $0.325$ & $7$ & $0.355$ & $13$ & $0.250$ \\
\underline{MS-COMET-QE-22}* & $13$ & $0.310$ & $6$ & $0.400$ & $13$ & $0.241$ \\
embed\_llama & $14$ & $0.250$ & $10$ & $0.242$ & $14$ & $0.180$ \\
XLsim & $14$ & $0.239$ & $6$ & $0.372$ & $16$ & $0.151$ \\
\underline{f200spBLEU} & $14$ & $0.237$ & $7$ & $0.343$ & $14$ & $0.178$ \\
\underline{chrF} & $15$ & $0.232$ & $8$ & $0.336$ & $15$ & $0.157$ \\
tokengram\_F & $16$ & $0.227$ & $8$ & $0.340$ & $16$ & $0.153$ \\
MEE4 & $17$ & $0.202$ & $7$ & $0.360$ & $16$ & $0.145$ \\
\underline{BLEU} & $17$ & $0.192$ & $9$ & $0.310$ & $17$ & $0.140$ \\
mre-score-labse-regular & $18$ & $0.111$ & $6$ & $0.376$ & $18$ & $0.087$ \\
\underline{Random-sysname}* & $19$ & $0.064$ & $11$ & $0.124$ & $19$ & $-0.015$ \\
eBLEU & $20$ & $-0.011$ & $8$ & $0.317$ & $19$ & $-0.030$ \\
\bottomrule
\end{tabular}

\caption{Segment-level Pearson correlation for the primary submissions to the WMT23 Metrics Shared Task, with sentinel metrics. The language direction is \langpair{en}{de}. Starred metrics are reference-free, underlined metrics are baselines, and highlighted metrics are sentinels. Ranks represent clusters of statistical significance and are computed following \citet{freitag-etal-2023-results}, which leverage the PERM-BOTH hypothesis test introduced by \citet{deutsch-etal-2021-statistical}.}
\label{tab:full-grouping-ende}
\end{table*}

%% file: tables/full-pearson-grouping-heen.tex
\begin{table*}[ht]
\small
\centering
\begin{tabular}{lrrrrrr}
\toprule
\multicolumn{1}{l}{\textbf{Metric}} & \multicolumn{2}{c}{\textbf{No}} & \multicolumn{2}{c}{\textbf{Segment}} & \multicolumn{2}{c}{\textbf{System}} \\

\cmidrule{2-7}
XCOMET-Ensemble & $1$ & $0.556$ & $1$ & $0.479$ & $1$ & $0.515$ \\
MetricX-23 & $1$ & $0.548$ & $2$ & $0.441$ & $1$ & $0.509$ \\
MetricX-23-QE* & $2$ & $0.520$ & $5$ & $0.387$ & $2$ & $0.480$ \\
XCOMET-QE-Ensemble* & $3$ & $0.498$ & $4$ & $0.397$ & $3$ & $0.458$ \\
MaTESe & $4$ & $0.459$ & $5$ & $0.373$ & $4$ & $0.408$ \\
mbr-metricx-qe* & $5$ & $0.411$ & $2$ & $0.448$ & $5$ & $0.362$ \\
GEMBA-MQM* & $5$ & $0.401$ & $2$ & $0.431$ & $6$ & $0.354$ \\
\underline{COMET} & $5$ & $0.401$ & $3$ & $0.421$ & $5$ & $0.367$ \\
Calibri-COMET22 & $5$ & $0.401$ & $4$ & $0.397$ & $5$ & $0.371$ \\
\underline{YiSi-1} & $6$ & $0.395$ & $2$ & $0.439$ & $6$ & $0.348$ \\
Calibri-COMET22-QE* & $6$ & $0.395$ & $6$ & $0.354$ & $5$ & $0.369$ \\
\underline{CometKiwi}* & $7$ & $0.387$ & $5$ & $0.375$ & $6$ & $0.353$ \\
sescoreX & $7$ & $0.385$ & $5$ & $0.370$ & $6$ & $0.352$ \\
KG-BERTScore* & $8$ & $0.382$ & $5$ & $0.375$ & $7$ & $0.347$ \\
\underline{BLEURT-20} & $8$ & $0.382$ & $3$ & $0.418$ & $7$ & $0.344$ \\
mre-score-labse-regular & $8$ & $0.378$ & $4$ & $0.407$ & $8$ & $0.335$ \\
cometoid22-wmt22* & $9$ & $0.365$ & $7$ & $0.309$ & $7$ & $0.346$ \\
\underline{docWMT22CometDA} & $10$ & $0.339$ & $5$ & $0.379$ & $9$ & $0.294$ \\
\rowcolor{lightred}
\candonlyready* & $10$ & $0.339$ & $11$ & $0.104$ & $7$ & $0.343$ \\
\underline{BERTscore} & $10$ & $0.335$ & $4$ & $0.412$ & $9$ & $0.293$ \\
\rowcolor{lightred}
\srconlyready* & $10$ & $0.334$ & $13$ & $0.000$ & $7$ & $0.336$ \\
\underline{prismRef} & $11$ & $0.319$ & $3$ & $0.428$ & $10$ & $0.276$ \\
\rowcolor{lightred}
\refonly & $11$ & $0.301$ & $13$ & $0.000$ & $9$ & $0.299$ \\
\underline{MS-COMET-QE-22}* & $12$ & $0.295$ & $9$ & $0.252$ & $10$ & $0.274$ \\
\underline{docWMT22CometKiwiDA}* & $12$ & $0.286$ & $7$ & $0.324$ & $11$ & $0.234$ \\
MEE4 & $13$ & $0.256$ & $8$ & $0.291$ & $11$ & $0.222$ \\
XLsim & $14$ & $0.233$ & $7$ & $0.314$ & $12$ & $0.198$ \\
\underline{f200spBLEU} & $14$ & $0.230$ & $8$ & $0.287$ & $12$ & $0.195$ \\
tokengram\_F & $14$ & $0.226$ & $7$ & $0.311$ & $13$ & $0.184$ \\
\underline{chrF} & $15$ & $0.221$ & $7$ & $0.308$ & $14$ & $0.179$ \\
\underline{BLEU} & $15$ & $0.220$ & $9$ & $0.260$ & $13$ & $0.189$ \\
embed\_llama & $15$ & $0.215$ & $10$ & $0.188$ & $13$ & $0.187$ \\
\underline{prismSrc}* & $16$ & $0.140$ & $11$ & $0.100$ & $15$ & $0.150$ \\
eBLEU & $16$ & $0.131$ & $8$ & $0.280$ & $16$ & $0.104$ \\
\underline{Random-sysname}* & $17$ & $0.041$ & $12$ & $0.057$ & $17$ & $0.001$ \\
\bottomrule
\end{tabular}

\caption{Segment-level Pearson correlation for the primary submissions to the WMT23 Metrics Shared Task, with sentinel metrics. The language direction is \langpair{he}{en}. Starred metrics are reference-free, underlined metrics are baselines, and highlighted metrics are sentinels. Ranks represent clusters of statistical significance and are computed following \citet{freitag-etal-2023-results}, which leverage the PERM-BOTH hypothesis test introduced by \citet{deutsch-etal-2021-statistical}.}
\label{tab:full-grouping-heen}
\end{table*}

%% file: tables/full-kendall-grouping-zhen.tex
\begin{table*}[ht]
\centering
\small
\begin{tabular}{lrrrrrr}
\toprule
\multicolumn{1}{l}{\textbf{Metric}} & \multicolumn{2}{c}{\textbf{No}} & \multicolumn{2}{c}{\textbf{Segment}}  & \multicolumn{2}{c}{\textbf{System}} \\
\cmidrule{2-7}
XCOMET-Ensemble & $1$ & $0.473$ & $2$ & $0.299$ & $1$ & $0.456$ \\
XCOMET-QE-Ensemble* & $2$ & $0.467$ & $3$ & $0.273$ & $2$ & $0.451$ \\
MetricX-23-QE* & $3$ & $0.461$ & $4$ & $0.252$ & $2$ & $0.448$ \\
GEMBA-MQM* & $3$ & $0.457$ & $1$ & $0.365$ & $4$ & $0.416$ \\
MetricX-23 & $4$ & $0.449$ & $3$ & $0.269$ & $3$ & $0.434$ \\
mbr-metricx-qe* & $5$ & $0.427$ & $2$ & $0.301$ & $5$ & $0.403$ \\
cometoid22-wmt22* & $5$ & $0.423$ & $4$ & $0.252$ & $4$ & $0.408$ \\
\rowcolor{lightred}
\candonlyready* & $6$ & $0.404$ & $9$ & $0.148$ & $4$ & $0.410$ \\
\rowcolor{lightred}
\srconlyready* & $7$ & $0.397$ & $14$ & $0.000$ & $4$ & $0.411$ \\
\underline{CometKiwi}* & $7$ & $0.391$ & $3$ & $0.263$ & $6$ & $0.368$ \\
Calibri-COMET22-QE* & $8$ & $0.386$ & $4$ & $0.241$ & $6$ & $0.366$ \\
sescoreX & $9$ & $0.375$ & $6$ & $0.217$ & $6$ & $0.367$ \\
MaTESe & $9$ & $0.371$ & $3$ & $0.271$ & $7$ & $0.345$ \\
KG-BERTScore* & $10$ & $0.361$ & $4$ & $0.248$ & $8$ & $0.337$ \\
\rowcolor{lightred}
\refonly & $11$ & $0.340$ & $14$ & $0.000$ & $7$ & $0.353$ \\
\underline{COMET} & $11$ & $0.333$ & $4$ & $0.248$ & $9$ & $0.311$ \\
\underline{MS-COMET-QE-22}* & $11$ & $0.332$ & $6$ & $0.213$ & $9$ & $0.311$ \\
Calibri-COMET22 & $12$ & $0.330$ & $6$ & $0.217$ & $9$ & $0.310$ \\
\underline{BLEURT-20} & $13$ & $0.310$ & $3$ & $0.261$ & $10$ & $0.288$ \\
\underline{docWMT22CometKiwiDA}* & $14$ & $0.299$ & $5$ & $0.234$ & $11$ & $0.265$ \\
\underline{docWMT22CometDA} & $15$ & $0.276$ & $5$ & $0.231$ & $12$ & $0.248$ \\
\underline{prismSrc}* & $16$ & $0.234$ & $12$ & $0.044$ & $12$ & $0.251$ \\
\underline{YiSi-1} & $17$ & $0.220$ & $5$ & $0.231$ & $13$ & $0.196$ \\
\underline{BERTscore} & $18$ & $0.180$ & $6$ & $0.216$ & $14$ & $0.156$ \\
mre-score-labse-regular & $18$ & $0.178$ & $7$ & $0.176$ & $14$ & $0.165$ \\
\underline{prismRef} & $19$ & $0.165$ & $5$ & $0.232$ & $15$ & $0.140$ \\
embed\_llama & $20$ & $0.109$ & $11$ & $0.096$ & $16$ & $0.093$ \\
XLsim & $20$ & $0.101$ & $10$ & $0.140$ & $17$ & $0.080$ \\
MEE4 & $21$ & $0.091$ & $8$ & $0.172$ & $18$ & $0.064$ \\
\underline{BLEU} & $21$ & $0.085$ & $9$ & $0.154$ & $18$ & $0.062$ \\
\underline{f200spBLEU} & $22$ & $0.068$ & $8$ & $0.165$ & $19$ & $0.042$ \\
\underline{chrF} & $23$ & $0.045$ & $7$ & $0.187$ & $20$ & $0.017$ \\
tokengram\_F & $24$ & $0.042$ & $7$ & $0.187$ & $21$ & $0.012$ \\
\underline{Random-sysname}* & $25$ & $0.015$ & $13$ & $0.025$ & $22$ & $-0.005$ \\
eBLEU & $26$ & $-0.041$ & $9$ & $0.156$ & $23$ & $-0.064$ \\
\bottomrule
\end{tabular}

\caption{Segment-level Kendall $\tau$ correlation coefficient for the primary submissions to the WMT23 Metrics Shared Task, with sentinel metrics. The language direction is \langpair{zh}{en}. Starred metrics are reference-free, underlined metrics are baselines, and highlighted metrics are sentinels. Ranks represent clusters of statistical significance and are computed following \citet{freitag-etal-2023-results}, which leverage the PERM-BOTH hypothesis test introduced by \citet{deutsch-etal-2021-statistical}.}
\label{tab:full-kendall-grouping-zhen}
\end{table*}

%% file: tables/full-kendall-grouping-ende.tex
\begin{table*}[ht]
\centering
\small
\begin{tabular}{lrrrrrr}
\toprule
\multicolumn{1}{l}{\textbf{Metric}} & \multicolumn{2}{c}{\textbf{No}} & \multicolumn{2}{c}{\textbf{Segment}}  & \multicolumn{2}{c}{\textbf{System}} \\
\cmidrule{2-7}
XCOMET-Ensemble & $1$ & $0.546$ & $1$ & $0.380$ & $1$ & $0.530$ \\
XCOMET-QE-Ensemble* & $2$ & $0.532$ & $2$ & $0.360$ & $2$ & $0.516$ \\
MetricX-23-QE* & $3$ & $0.509$ & $2$ & $0.357$ & $3$ & $0.487$ \\
MetricX-23 & $3$ & $0.506$ & $2$ & $0.368$ & $3$ & $0.485$ \\
sescoreX & $4$ & $0.493$ & $3$ & $0.343$ & $4$ & $0.476$ \\
mbr-metricx-qe* & $4$ & $0.490$ & $1$ & $0.397$ & $4$ & $0.467$ \\
GEMBA-MQM* & $4$ & $0.482$ & $1$ & $0.399$ & $5$ & $0.449$ \\
\rowcolor{lightred}
\candonlyready* & $5$ & $0.463$ & $4$ & $0.290$ & $5$ & $0.456$ \\
MaTESe & $5$ & $0.462$ & $5$ & $0.286$ & $6$ & $0.447$ \\
\underline{BLEURT-20} & $6$ & $0.452$ & $2$ & $0.366$ & $7$ & $0.426$ \\
\rowcolor{lightred}
\srconlyready* & $6$ & $0.443$ & $11$ & $0.000$ & $5$ & $0.462$ \\
cometoid22-wmt22* & $7$ & $0.422$ & $2$ & $0.362$ & $8$ & $0.398$ \\
\rowcolor{lightred}
\refonly & $7$ & $0.418$ & $11$ & $0.000$ & $6$ & $0.437$ \\
\underline{COMET} & $7$ & $0.418$ & $2$ & $0.366$ & $9$ & $0.387$ \\
Calibri-COMET22 & $7$ & $0.417$ & $3$ & $0.342$ & $9$ & $0.387$ \\
\underline{CometKiwi}* & $8$ & $0.408$ & $3$ & $0.330$ & $9$ & $0.379$ \\
Calibri-COMET22-QE* & $8$ & $0.406$ & $5$ & $0.279$ & $9$ & $0.379$ \\
\underline{MS-COMET-QE-22}* & $9$ & $0.391$ & $5$ & $0.280$ & $10$ & $0.363$ \\
KG-BERTScore* & $10$ & $0.361$ & $4$ & $0.310$ & $11$ & $0.329$ \\
\underline{docWMT22CometKiwiDA}* & $10$ & $0.358$ & $4$ & $0.316$ & $11$ & $0.329$ \\
\underline{prismRef} & $11$ & $0.345$ & $6$ & $0.247$ & $11$ & $0.332$ \\
\underline{docWMT22CometDA} & $11$ & $0.337$ & $2$ & $0.360$ & $12$ & $0.296$ \\
\underline{YiSi-1} & $12$ & $0.280$ & $4$ & $0.297$ & $13$ & $0.250$ \\
\underline{prismSrc}* & $12$ & $0.267$ & $10$ & $0.039$ & $12$ & $0.284$ \\
\underline{BERTscore} & $13$ & $0.253$ & $5$ & $0.260$ & $14$ & $0.224$ \\
MEE4 & $14$ & $0.225$ & $5$ & $0.271$ & $15$ & $0.190$ \\
XLsim & $14$ & $0.217$ & $6$ & $0.257$ & $15$ & $0.180$ \\
\underline{f200spBLEU} & $15$ & $0.187$ & $6$ & $0.255$ & $16$ & $0.151$ \\
\underline{chrF} & $15$ & $0.186$ & $6$ & $0.241$ & $16$ & $0.152$ \\
tokengram\_F & $16$ & $0.183$ & $6$ & $0.245$ & $17$ & $0.149$ \\
embed\_llama & $16$ & $0.182$ & $8$ & $0.163$ & $16$ & $0.150$ \\
\underline{BLEU} & $17$ & $0.137$ & $7$ & $0.231$ & $18$ & $0.103$ \\
eBLEU & $18$ & $0.096$ & $7$ & $0.230$ & $19$ & $0.070$ \\
mre-score-labse-regular & $18$ & $0.084$ & $5$ & $0.269$ & $19$ & $0.066$ \\
\underline{Random-sysname}* & $19$ & $0.033$ & $9$ & $0.081$ & $20$ & $-0.018$ \\
\bottomrule
\end{tabular}

\caption{Segment-level Kendall $\tau$ correlation coefficient for the primary submissions to the WMT23 Metrics Shared Task, with sentinel metrics. The language direction is \langpair{en}{de}. Starred metrics are reference-free, underlined metrics are baselines, and highlighted metrics are sentinels. Ranks represent clusters of statistical significance and are computed following \citet{freitag-etal-2023-results}, which leverage the PERM-BOTH hypothesis test introduced by \citet{deutsch-etal-2021-statistical}.}
\label{tab:full-kendall-grouping-ende}
\end{table*}

%% file: tables/full-kendall-grouping-heen.tex
\begin{table*}[ht]
\centering
\small
\begin{tabular}{lrrrrrr}
\toprule
\multicolumn{1}{l}{\textbf{Metric}} & \multicolumn{2}{c}{\textbf{No}} & \multicolumn{2}{c}{\textbf{Segment}}  & \multicolumn{2}{c}{\textbf{System}} \\
\cmidrule{2-7}
XCOMET-Ensemble & $1$ & $0.415$ & $2$ & $0.323$ & $1$ & $0.395$ \\
MetricX-23 & $2$ & $0.401$ & $3$ & $0.302$ & $2$ & $0.382$ \\
GEMBA-MQM* & $2$ & $0.399$ & $1$ & $0.369$ & $3$ & $0.367$ \\
XCOMET-QE-Ensemble* & $3$ & $0.374$ & $5$ & $0.276$ & $3$ & $0.358$ \\
MetricX-23-QE* & $3$ & $0.370$ & $6$ & $0.251$ & $3$ & $0.355$ \\
mbr-metricx-qe* & $3$ & $0.366$ & $2$ & $0.316$ & $4$ & $0.339$ \\
MaTESe & $4$ & $0.361$ & $3$ & $0.302$ & $4$ & $0.341$ \\
\underline{COMET} & $5$ & $0.350$ & $3$ & $0.309$ & $5$ & $0.327$ \\
Calibri-COMET22 & $6$ & $0.348$ & $4$ & $0.284$ & $6$ & $0.324$ \\
\underline{BLEURT-20} & $6$ & $0.344$ & $4$ & $0.295$ & $6$ & $0.320$ \\
sescoreX & $6$ & $0.342$ & $4$ & $0.285$ & $6$ & $0.320$ \\
\underline{CometKiwi}* & $7$ & $0.338$ & $6$ & $0.238$ & $6$ & $0.323$ \\
Calibri-COMET22-QE* & $7$ & $0.336$ & $7$ & $0.230$ & $6$ & $0.322$ \\
\underline{YiSi-1} & $7$ & $0.333$ & $2$ & $0.325$ & $7$ & $0.303$ \\
mre-score-labse-regular & $7$ & $0.328$ & $4$ & $0.284$ & $7$ & $0.300$ \\
KG-BERTScore* & $8$ & $0.322$ & $6$ & $0.242$ & $7$ & $0.304$ \\
cometoid22-wmt22* & $9$ & $0.310$ & $7$ & $0.216$ & $7$ & $0.301$ \\
\underline{prismRef} & $9$ & $0.302$ & $3$ & $0.309$ & $8$ & $0.273$ \\
\underline{BERTscore} & $10$ & $0.295$ & $4$ & $0.298$ & $9$ & $0.266$ \\
\underline{docWMT22CometDA} & $11$ & $0.278$ & $5$ & $0.270$ & $10$ & $0.249$ \\
\underline{MS-COMET-QE-22}* & $12$ & $0.261$ & $9$ & $0.174$ & $10$ & $0.249$ \\
\rowcolor{lightred}
\srconlyready* & $13$ & $0.243$ & $12$ & $0.000$ & $10$ & $0.247$ \\
\rowcolor{lightred}
\candonlyready* & $13$ & $0.243$ & $11$ & $0.049$ & $10$ & $0.249$ \\
XLsim & $13$ & $0.233$ & $7$ & $0.228$ & $11$ & $0.211$ \\
MEE4 & $13$ & $0.231$ & $7$ & $0.221$ & $11$ & $0.202$ \\
\underline{docWMT22CometKiwiDA}* & $14$ & $0.227$ & $7$ & $0.229$ & $12$ & $0.192$ \\
\rowcolor{lightred}
\refonly & $15$ & $0.210$ & $12$ & $0.000$ & $11$ & $0.214$ \\
tokengram\_F & $15$ & $0.207$ & $7$ & $0.228$ & $13$ & $0.175$ \\
\underline{chrF} & $16$ & $0.204$ & $7$ & $0.224$ & $14$ & $0.171$ \\
\underline{f200spBLEU} & $17$ & $0.193$ & $7$ & $0.219$ & $15$ & $0.162$ \\
\underline{BLEU} & $18$ & $0.184$ & $8$ & $0.205$ & $16$ & $0.157$ \\
embed\_llama & $18$ & $0.174$ & $10$ & $0.147$ & $16$ & $0.151$ \\
eBLEU & $19$ & $0.166$ & $8$ & $0.209$ & $17$ & $0.141$ \\
\underline{prismSrc}* & $19$ & $0.164$ & $11$ & $0.043$ & $14$ & $0.169$ \\
\underline{Random-sysname}* & $20$ & $0.027$ & $11$ & $0.033$ & $18$ & $0.002$ \\
\bottomrule
\end{tabular}

\caption{Segment-level Kendall $\tau$ correlation coefficient for the primary submissions to the WMT23 Metrics Shared Task, with sentinel metrics. The language direction is \langpair{he}{en}. Starred metrics are reference-free, underlined metrics are baselines, and highlighted metrics are sentinels. Ranks represent clusters of statistical significance and are computed following \citet{freitag-etal-2023-results}, which leverage the PERM-BOTH hypothesis test introduced by \citet{deutsch-etal-2021-statistical}.}
\label{tab:full-kendall-grouping-heen}
\end{table*}

%% file: figures/correlations-zhen.tex
\begin{figure*}
    \centering
    \resizebox{\textwidth}{!}{
    \includegraphics{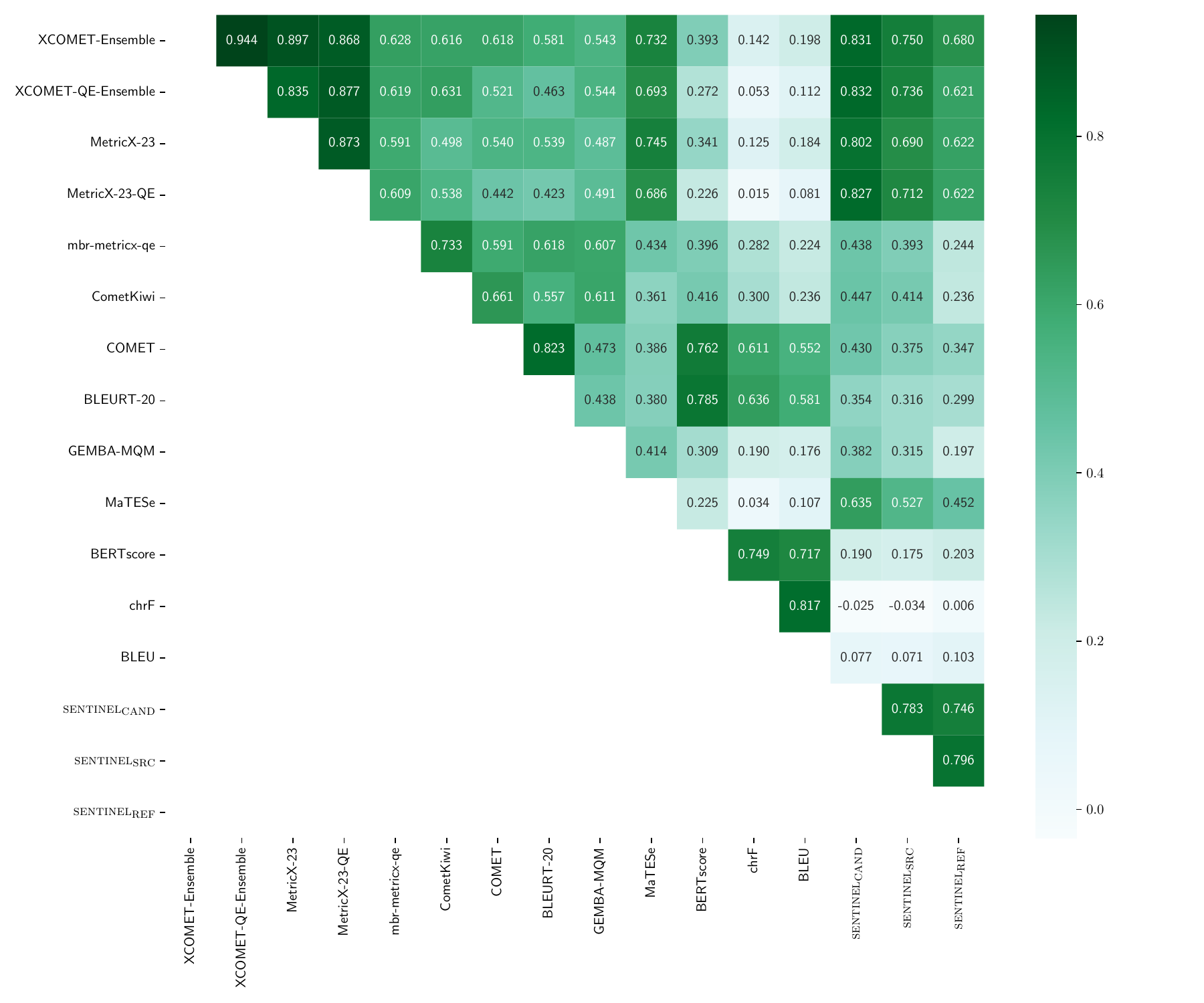}
    }
    \caption{Pairwise correlation between a part of the primary submissions and baselines of WMT23, and sentinel metrics. Correlation is Pearson with \textit{No Grouping}, and the language direction is \langpair{zh}{en}.}
    \label{fig:correlations-zhen}
\end{figure*}

%% file: figures/correlations-ende.tex
\begin{figure*}
    \centering
    \resizebox{\textwidth}{!}{
    \includegraphics{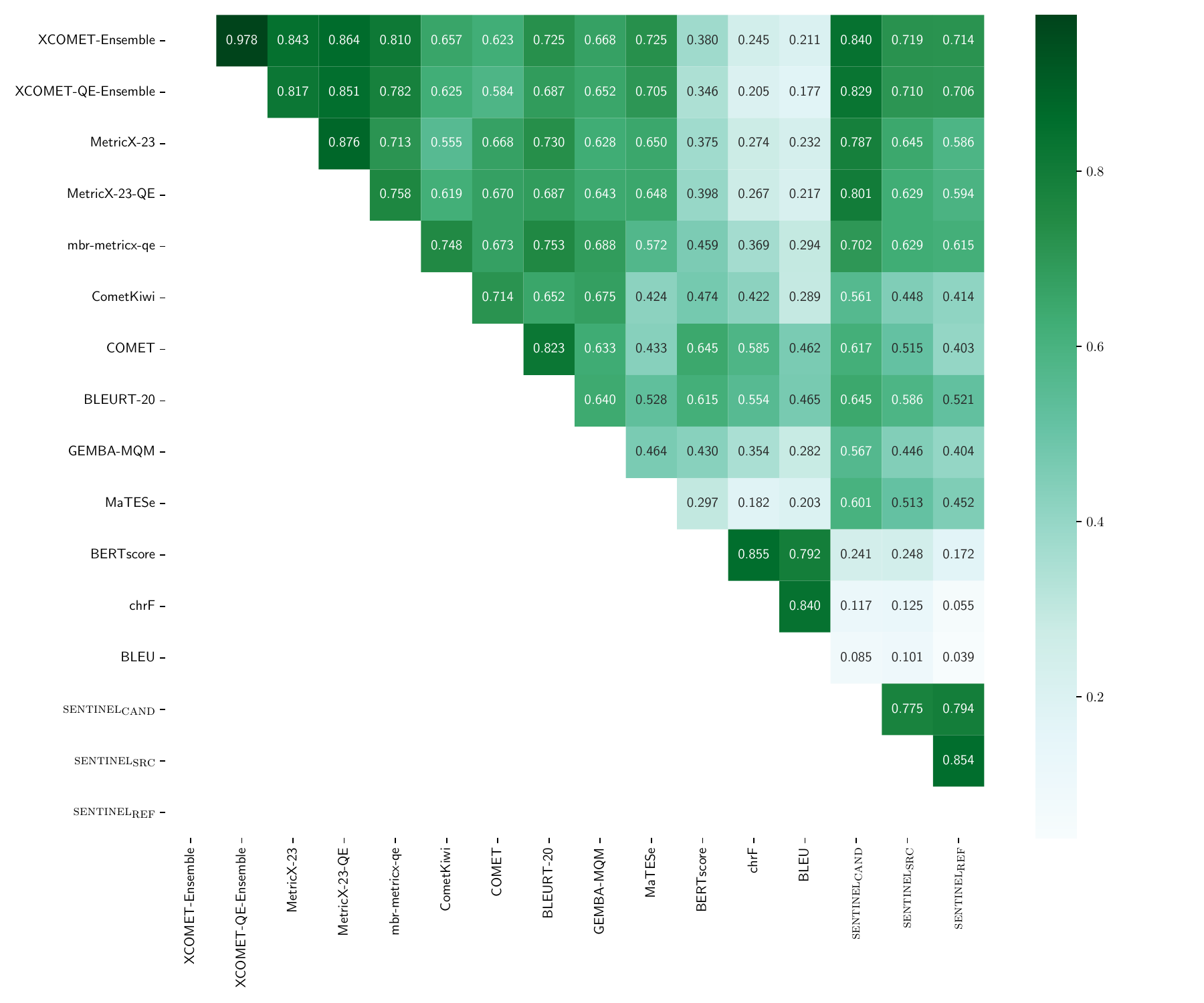}
    }
    \caption{Pairwise correlation between a part of the primary submissions and baselines of WMT23, and sentinel metrics. Correlation is Pearson with \textit{No Grouping}, and the language direction is \langpair{en}{de}.}
    \label{fig:correlations-ende}
\end{figure*}

%% file: figures/correlations-heen.tex
\begin{figure*}
    \centering
    \resizebox{\textwidth}{!}{
    \includegraphics{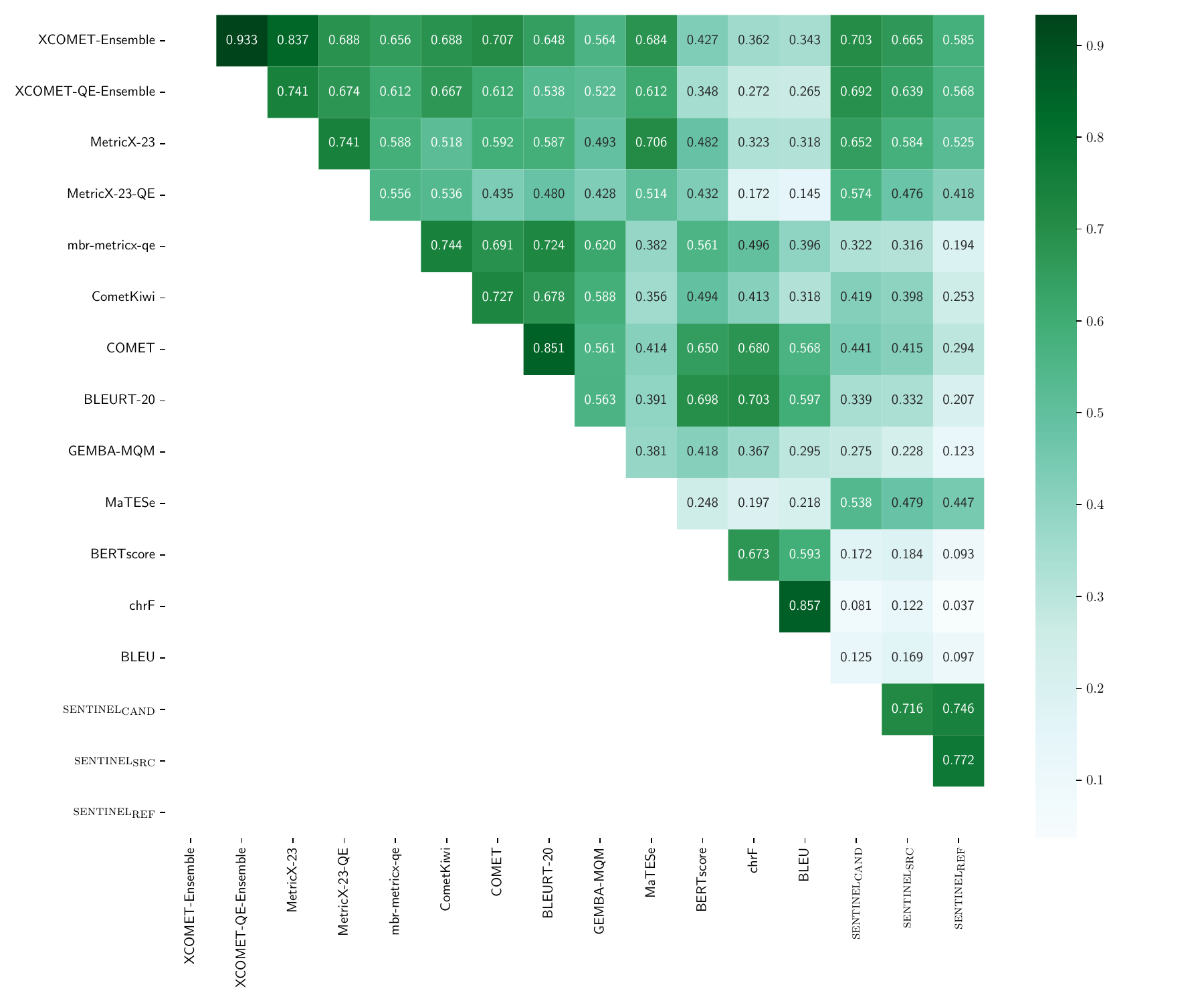}
    }
    \caption{Pairwise correlation between a part of the primary submissions and baselines of WMT23, and sentinel metrics. Correlation is Pearson with \textit{No Grouping}, and the language direction is \langpair{he}{en}.}
    \label{fig:correlations-heen}
\end{figure*}

%% file: figures/length-bias/all-1.tex
\begin{figure*}[!ht]
    \centering
    \begin{subfigure}[b]{0.9\columnwidth}
        \includegraphics[width=\textwidth]{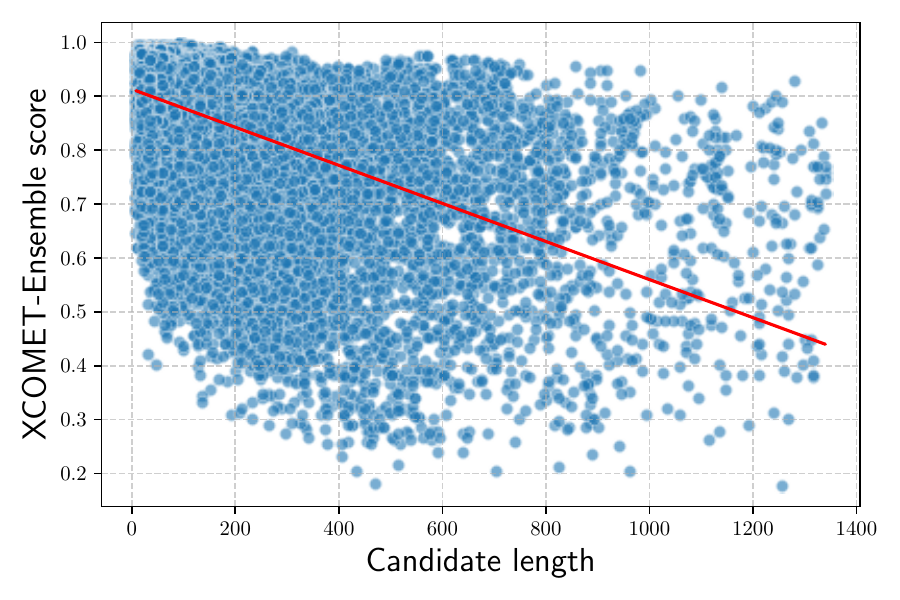}
    \end{subfigure}
    \begin{subfigure}[b]{0.9\columnwidth}
        \includegraphics[width=\textwidth]{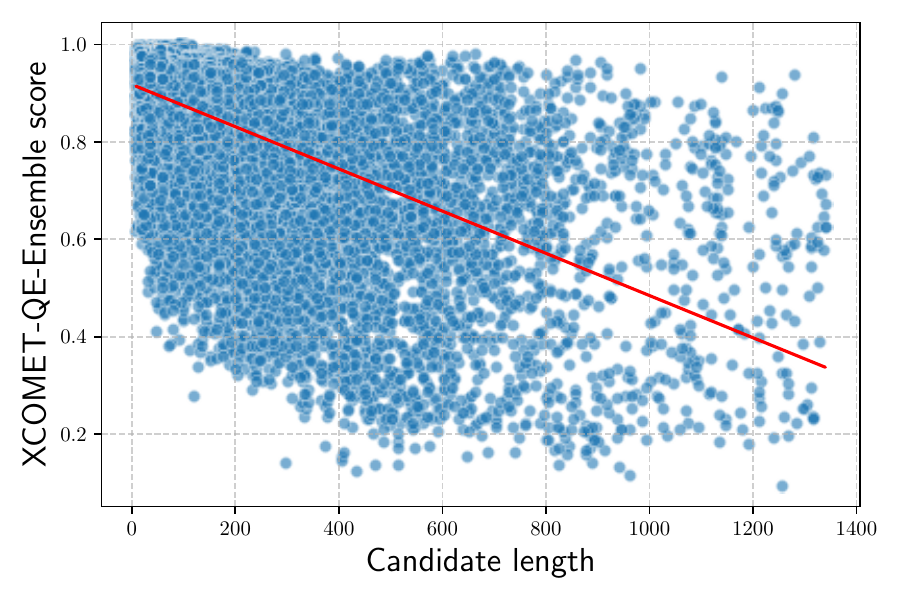}
    \end{subfigure}
    \begin{subfigure}[b]{0.9\columnwidth}
        \includegraphics[width=\textwidth]{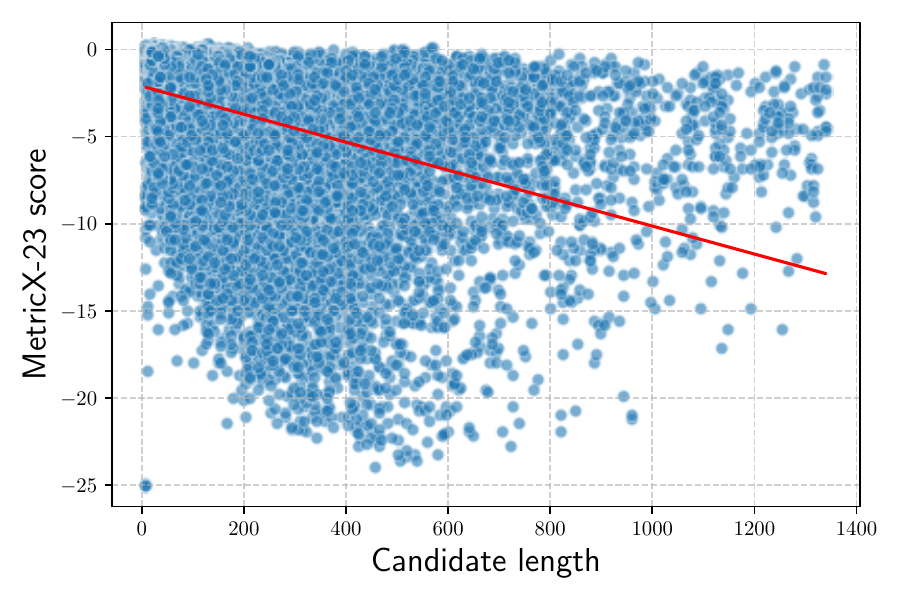}
    \end{subfigure}
    \begin{subfigure}[b]{0.9\columnwidth}
        \includegraphics[width=\textwidth]{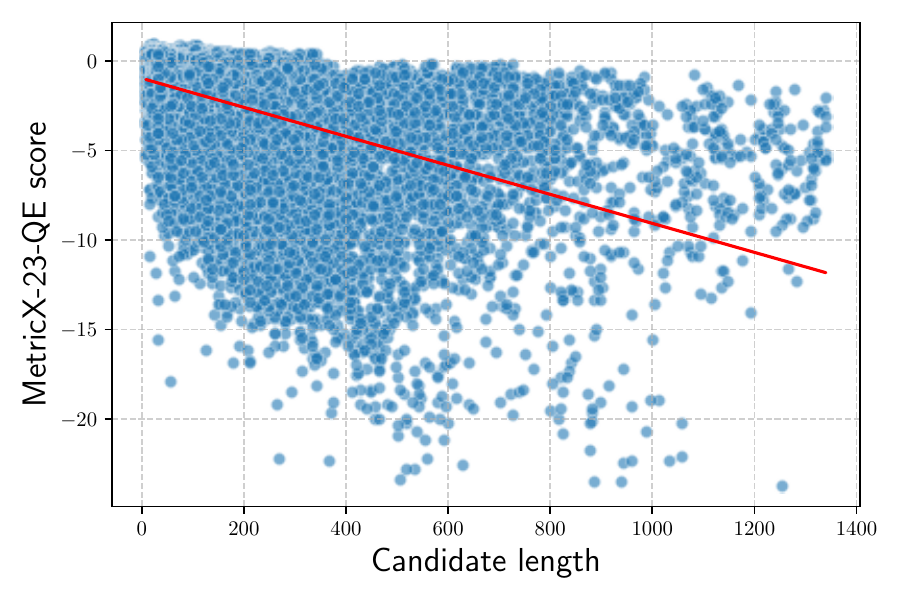}
    \end{subfigure}
    \begin{subfigure}[b]{0.9\columnwidth}
        \includegraphics[width=\textwidth]{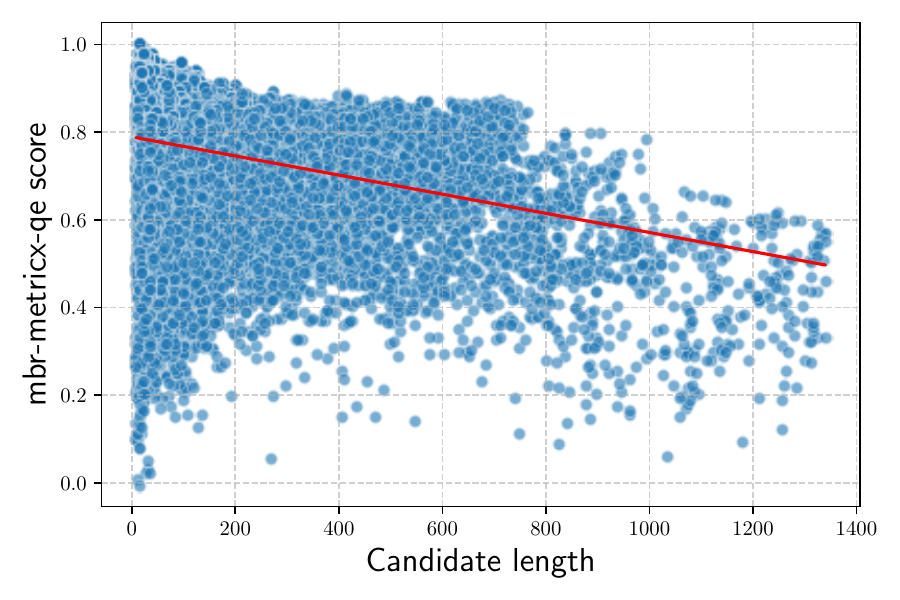}
    \end{subfigure}
    \begin{subfigure}[b]{0.9\columnwidth}
        \includegraphics[width=\textwidth]{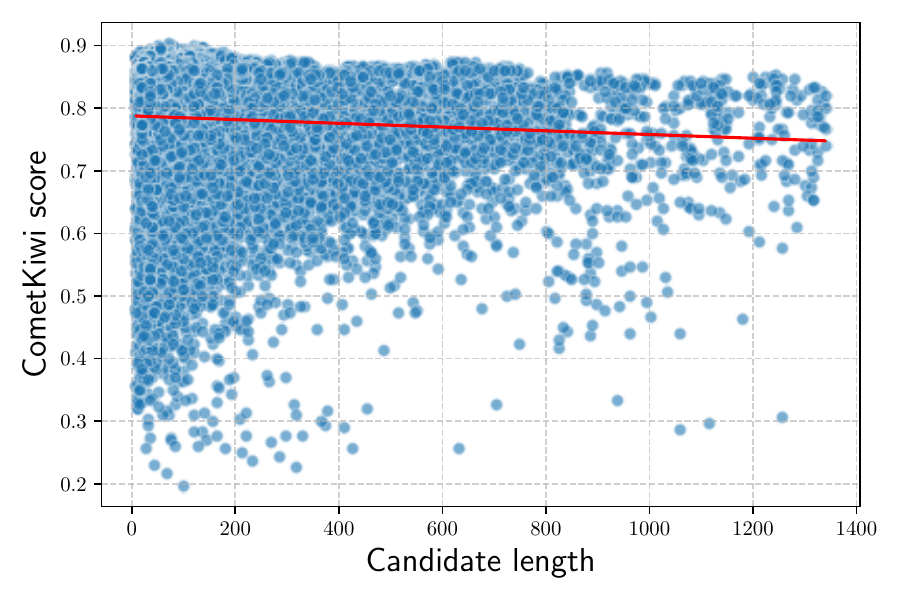}
    \end{subfigure}
    \begin{subfigure}[b]{0.9\columnwidth}
        \includegraphics[width=\textwidth]{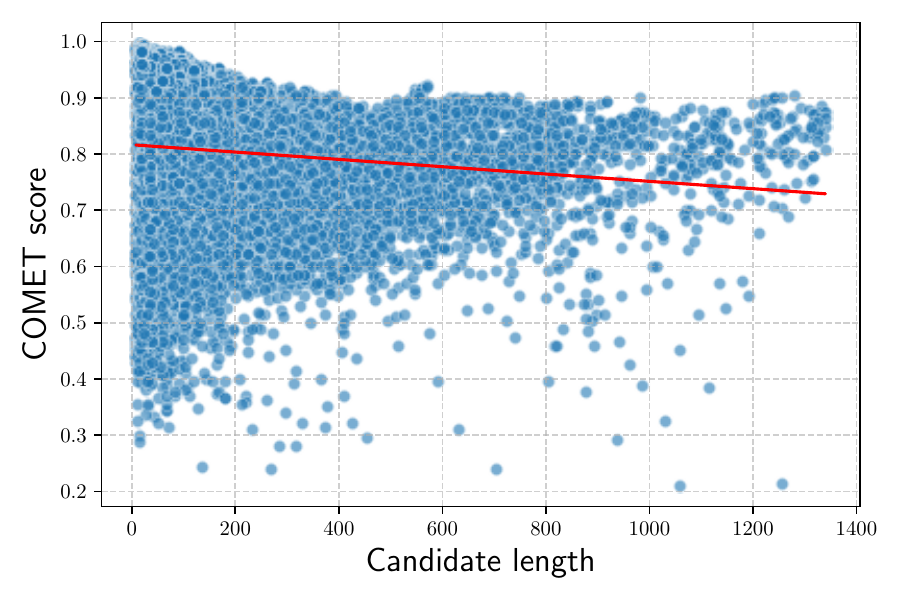}
    \end{subfigure}
    \begin{subfigure}[b]{0.9\columnwidth}
        \includegraphics[width=\textwidth]{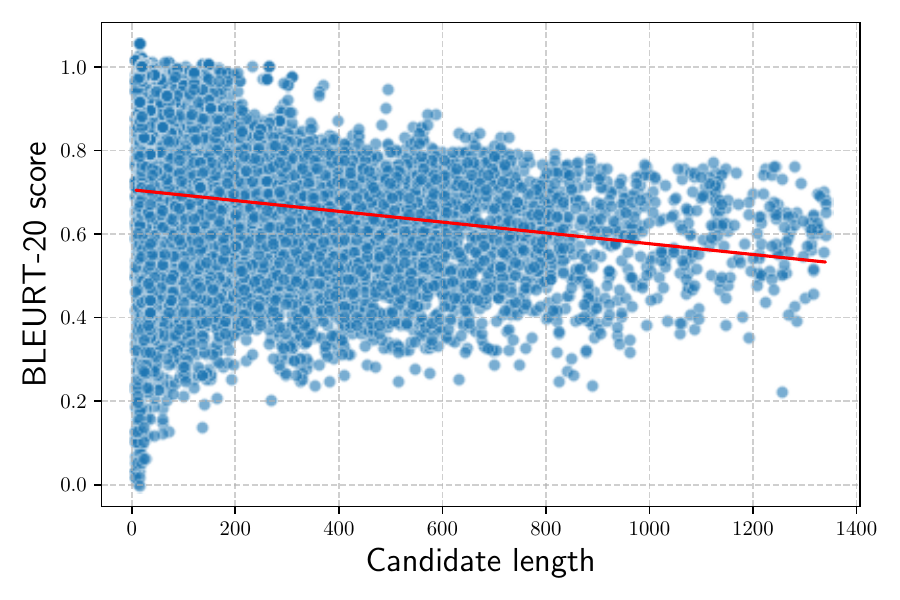}
    \end{subfigure}
    \begin{subfigure}[b]{0.9\columnwidth}
        \includegraphics[width=\textwidth]{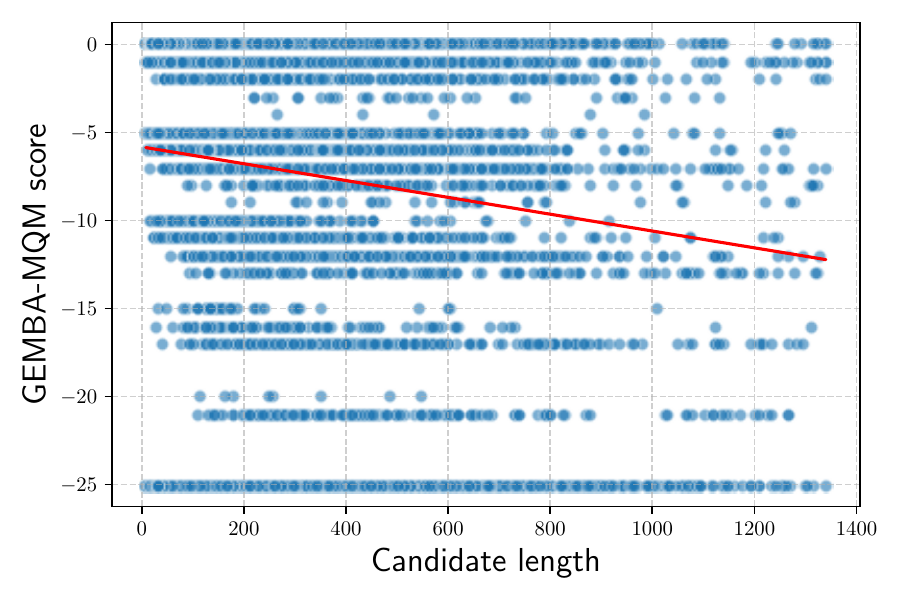}
    \end{subfigure}
    \begin{subfigure}[b]{0.9\columnwidth}
        \includegraphics[width=\textwidth]{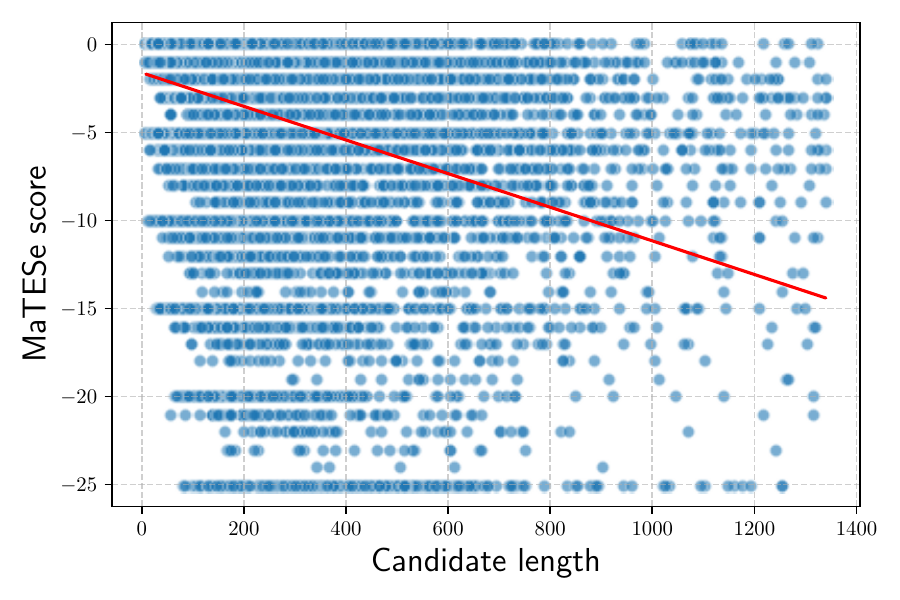}
    \end{subfigure}
    \caption{Metric assessments over translation length for a subset of the metrics that participated in WMT23. The red line represents the least-squares regression.}
    \label{fig:length-bias-all-1}
\end{figure*}

%% file: figures/length-bias/all-2.tex
\begin{figure*}[!ht]
    \centering
    \begin{subfigure}[b]{0.9\columnwidth}
        \includegraphics[width=\textwidth]{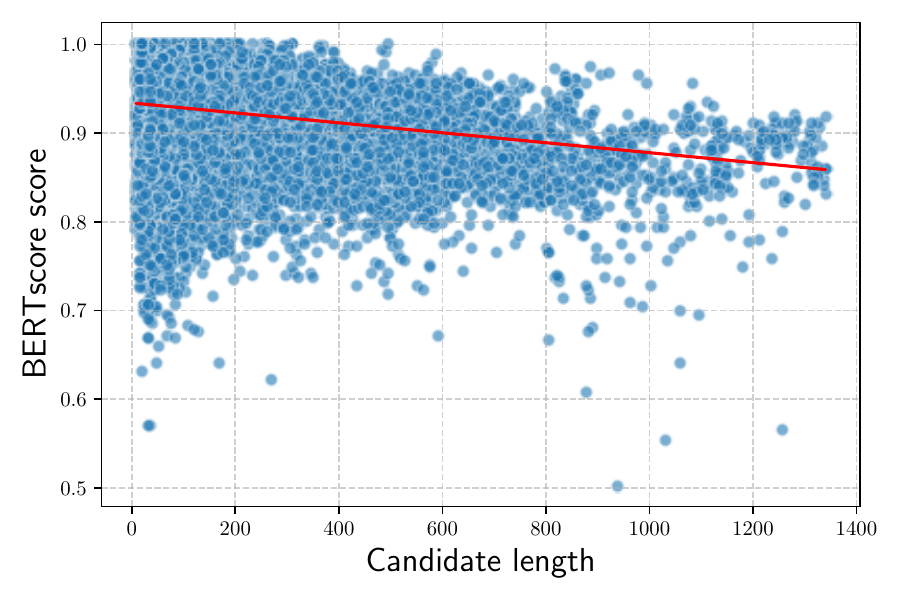}
    \end{subfigure}
    \begin{subfigure}[b]{0.9\columnwidth}
        \includegraphics[width=\textwidth]{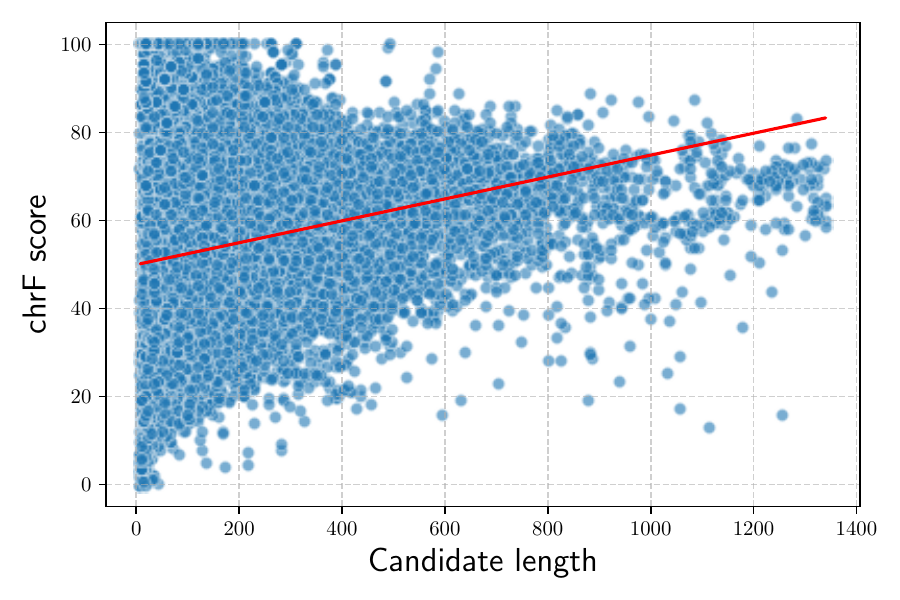}
    \end{subfigure}
    \begin{subfigure}[b]{0.9\columnwidth}
        \includegraphics[width=\textwidth]{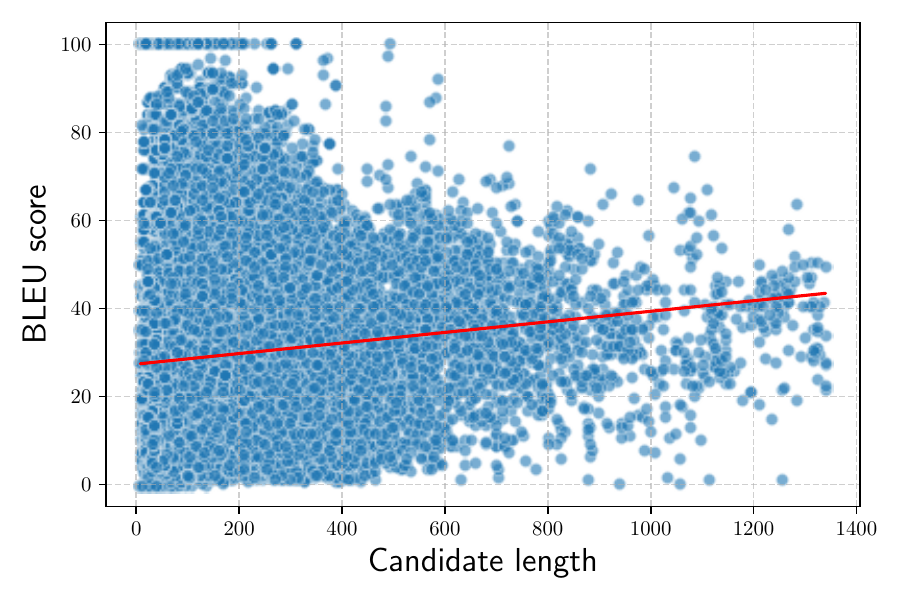}
    \end{subfigure}
    \begin{subfigure}[b]{0.9\columnwidth}
        \includegraphics[width=\textwidth]{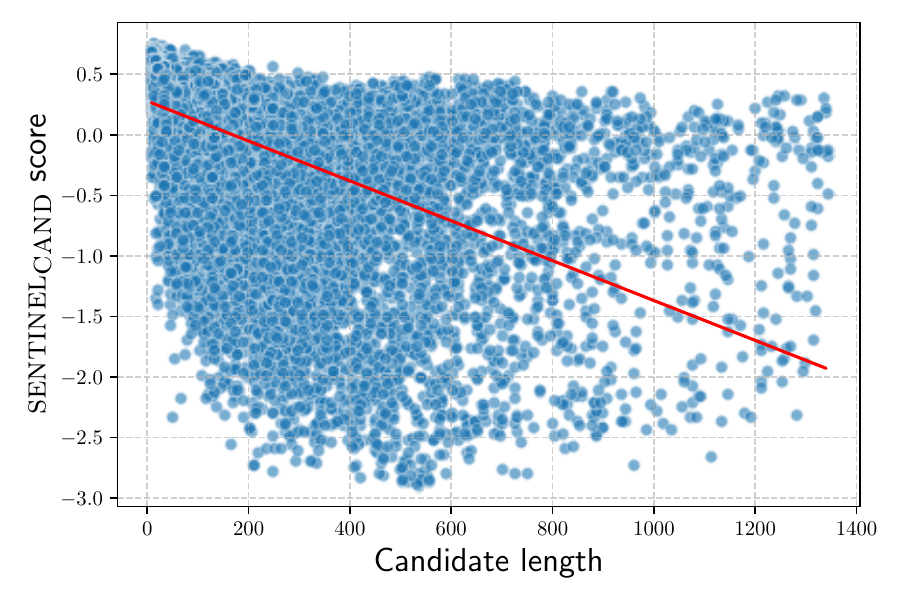}
    \end{subfigure}
    \begin{subfigure}[b]{0.9\columnwidth}
        \includegraphics[width=\textwidth]{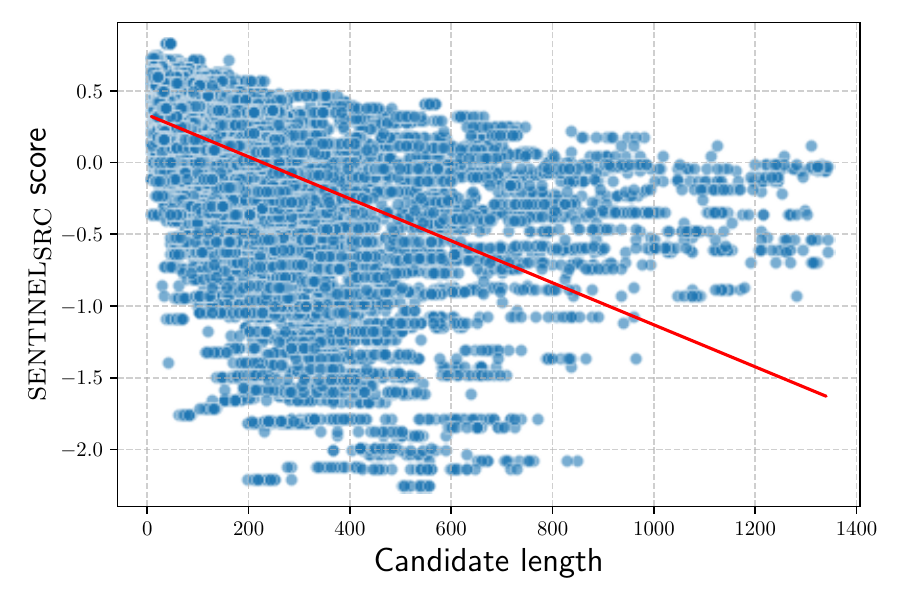}
    \end{subfigure}
    \begin{subfigure}[b]{0.9\columnwidth}
        \includegraphics[width=\textwidth]{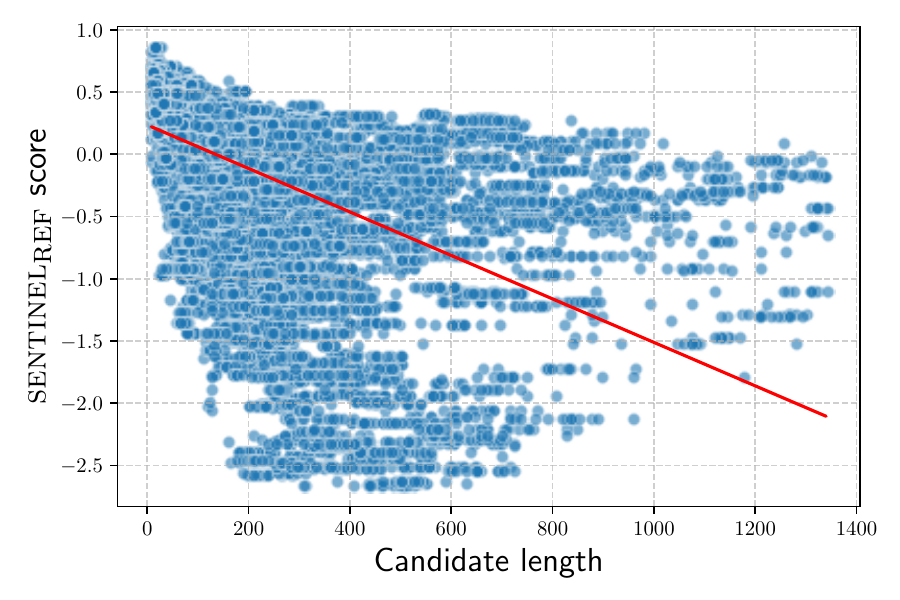}
    \end{subfigure}
    \begin{subfigure}[b]{0.9\columnwidth}
        \includegraphics[width=\textwidth]{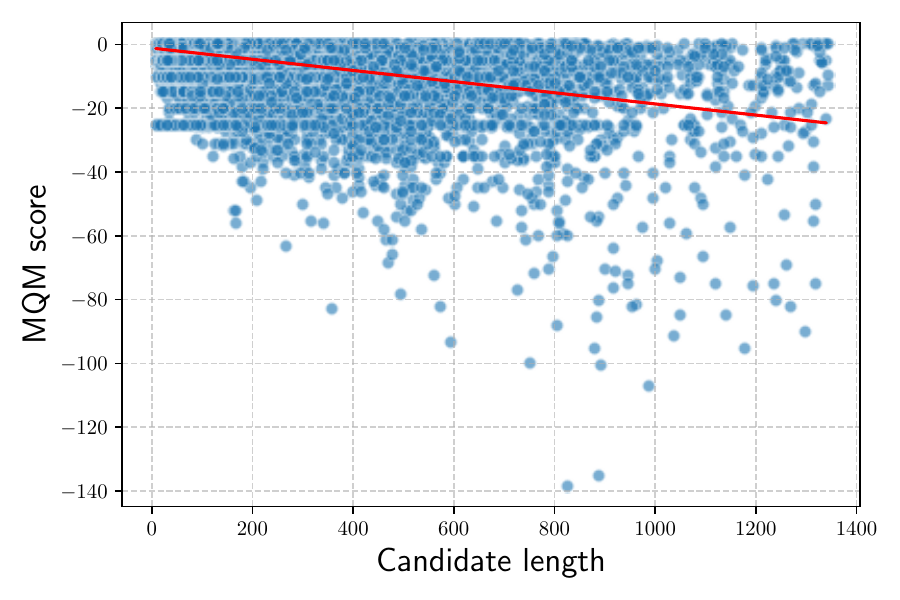}
    \end{subfigure}
    \caption{Metric assessments over translation length for a subset of the metrics that participated in WMT23, together with sentinel metrics. The red line represents the least-squares regression.}
    \label{fig:length-bias-all-2}
\end{figure*}

%% file: tables/ties-testsets.tex
\begin{table}[!ht]
    \small
    \centering
    \begin{tabular}{lrrrr}
        \toprule
                    &  \multicolumn{1}{c}{$2020$} & \multicolumn{1}{c}{$2021$} & \multicolumn{1}{c}{$2022$} & \multicolumn{1}{c}{$2023$}  \\ 
        \cmidrule{2-5}
        \langpair{en}{de}    & $15.14$ & $44.62$ & $53.35$ & $23.11$ \\
        \langpair{zh}{en}    & $17.01$ & $30.31$ & $41.55$ & $24.03$ \\
        \langpair{en}{ru}    & -- & $53.24$ & $44.42$ & -- \\
        \langpair{he}{en}    & -- & -- & -- & $42.84$ \\
    \end{tabular}
    \caption{Percentage of tied pairs in the MQM data released over different years at the Metrics Shared Task (or by \citet{freitag-etal-2021-experts}, for 2020), and regarding different translation directions.}
    \label{tab:ties-testsets}
\end{table}

%% file: tables/filtering-probs-zhen.tex
\begin{table*}[t]
\small
\centering
\resizebox{\textwidth}{!}{
\begin{tabular}{lrrrrrrrrrrrrr}
\toprule
$p_t$ & $1.00$ & $0.65$ & $0.30$ & $0.00$ & $0.00$ & $0.00$ & $0.00$ & $0.00$ & $0.00$ & $0.00$ & $0.00$ & $0.00$ & $0.00$ \\
$p_n$ & $0.00$ & $0.00$ & $0.00$ & $0.00$ & $0.20$ & $0.40$ & $0.50$ & $0.60$ & $0.65$ & $0.70$ & $0.75$ & $0.80$ & $0.85$ \\
$\%$ & $0$ & $10$ & $18$ & $24$ & $28$ & $35$ & $39$ & $44$ & $47$ & $51$ & $56$ & $61$ & $68$ \\
$\#$ & $93890$ & $104304$ & $114664$ & $123585$ & $104888$ & $85969$ & $76522$ & $67237$ & $62624$ & $57948$ & $53110$ & $48491$ & $43730$ \\
\bottomrule
\end{tabular}
}
\caption{$p_t$ is the probability of removing a tied human pair, and $p_n$ is that of removing a non-tied human pair. The considered test set is WMT23 \langpair{zh}{en}. Each column, i.e., each pair $(p_t, p_n)$, represents a sub-sample of the test set, in which tied and non-tied pairs have been removed with such probabilities. The third row contains the percentage of tied human pairs over all pairs, as a result of the sub-sampling. The last row contains the total number of pairs remaining in the test set after the sub-sampling.}
\label{tab:filtering-probs-zhen}
\end{table*}

%% file: tables/filtering-probs-ende.tex
\begin{table*}[t]
\small
\centering
\resizebox{\textwidth}{!}{
\begin{tabular}{lrrrrrrrrrrrrr}
\toprule
$p_t$ & $1.00$ & $0.65$ & $0.30$ & $0.00$ & $0.00$ & $0.00$ & $0.00$ & $0.00$ & $0.00$ & $0.00$ & $0.00$ & $0.00$ & $0.00$ \\
$p_n$ & $0.00$ & $0.00$ & $0.00$ & $0.00$ & $0.20$ & $0.40$ & $0.50$ & $0.60$ & $0.65$ & $0.7$ & $0.75$ & $0.80$ & $0.85$ \\
$\%$ & $0$ & $10$ & $17$ & $23$ & $27$ & $33$ & $38$ & $43$ & $46$ & $50$ & $54$ & $60$ & $67$ \\
$\#$ & $23343$ & $25803$ & $28236$ & $30360$ & $25694$ & $21021$ & $18689$ & $16353$ & $15184$ & $14014$ & $12899$ & $11698$ & $10493$ \\

\bottomrule
\end{tabular}
}
\caption{$p_t$ is the probability of removing a tied human pair, and $p_n$ is that of removing a non-tied human pair. The considered test set is WMT23 \langpair{en}{de}. Each column, i.e., each pair $(p_t, p_n)$, represents a sub-sample of the test set, in which tied and non-tied pairs have been removed with such probabilities. The third row contains the percentage of tied human pairs over all pairs, as a result of the sub-sampling. The last row contains the total number of pairs remaining in the test set after the sub-sampling.}
\label{tab:filtering-probs-ende}
\end{table*}

%% file: tables/filtering-probs-heen.tex
\begin{table*}[t]
\small
\centering
\resizebox{\textwidth}{!}{
\begin{tabular}{lrrrrrrrrrrrrr}
\toprule
$p_t$ & $1.0$ & $0.90$ & $0.80$ & $0.65$ & $0.50$ & $0.35$ & $0.20$ & $0.00$ & $0.00$ & $0.00$ & $0.00$ & $0.00$ & $0.00$ \\
$p_n$ & $0.00$ & $0.00$ & $0.00$ & $0.00$ & $0.00$ & $0.00$ & $0.00$ & $0.00$ & $0.20$ & $0.40$ & $0.55$ & $0.65$ & $0.75$ \\
$\%$ & $0$ & $7$ & $13$ & $21$ & $27$ & $33$ & $38$ & $43$ & $48$ & $56$ & $62$ & $68$ & $75$ \\
$\#$ & $36561$ & $39254$ & $42038$ & $46202$ & $50272$ & $54435$ & $58516$ & $63960$ & $56679$ & $49315$ & $43918$ & $40145$ & $36530$ \\
\bottomrule
\end{tabular}
}
\caption{$p_t$ is the probability of removing a tied human pair, and $p_n$ is that of removing a non-tied human pair. The considered test set is WMT23 \langpair{he}{en}.  Each column, i.e., each pair $(p_t, p_n)$, represents a sub-sample of the test set, in which tied and non-tied pairs have been removed with such probabilities. The third row contains the percentage of tied human pairs over all pairs, as a result of the sub-sampling. The last row contains the total number of pairs remaining in the test set after the sub-sampling.}
\label{tab:filtering-probs-heen}
\end{table*}

%% file: figures/accuracy-and-threshold-varying-ties-zhen.tex
\begin{figure*}[!ht]
    \centering
    \resizebox{\textwidth}{!}{
    \begin{subfigure}[b]{0.4\linewidth}
        \includegraphics[width=\linewidth]{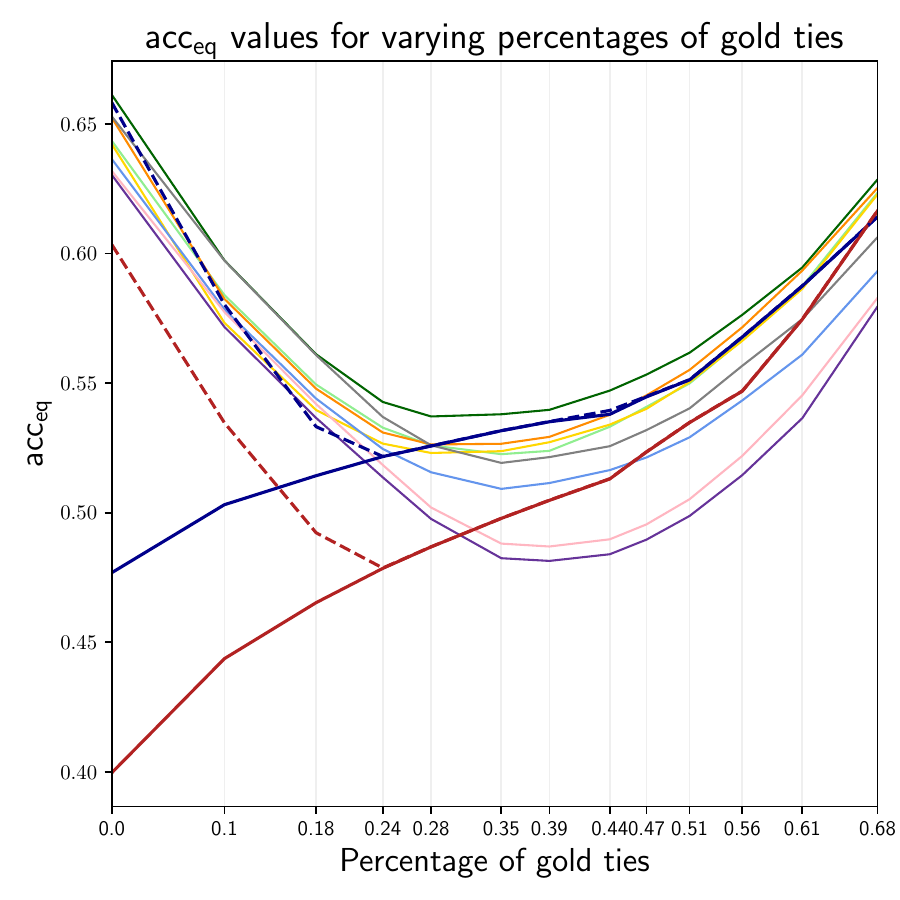}
        \caption{}
        \label{fig:accuracy-varying-ties-zhen}
    \end{subfigure}
    \quad
    \begin{subfigure}[b]{0.4\linewidth}
        \includegraphics[width=\linewidth]{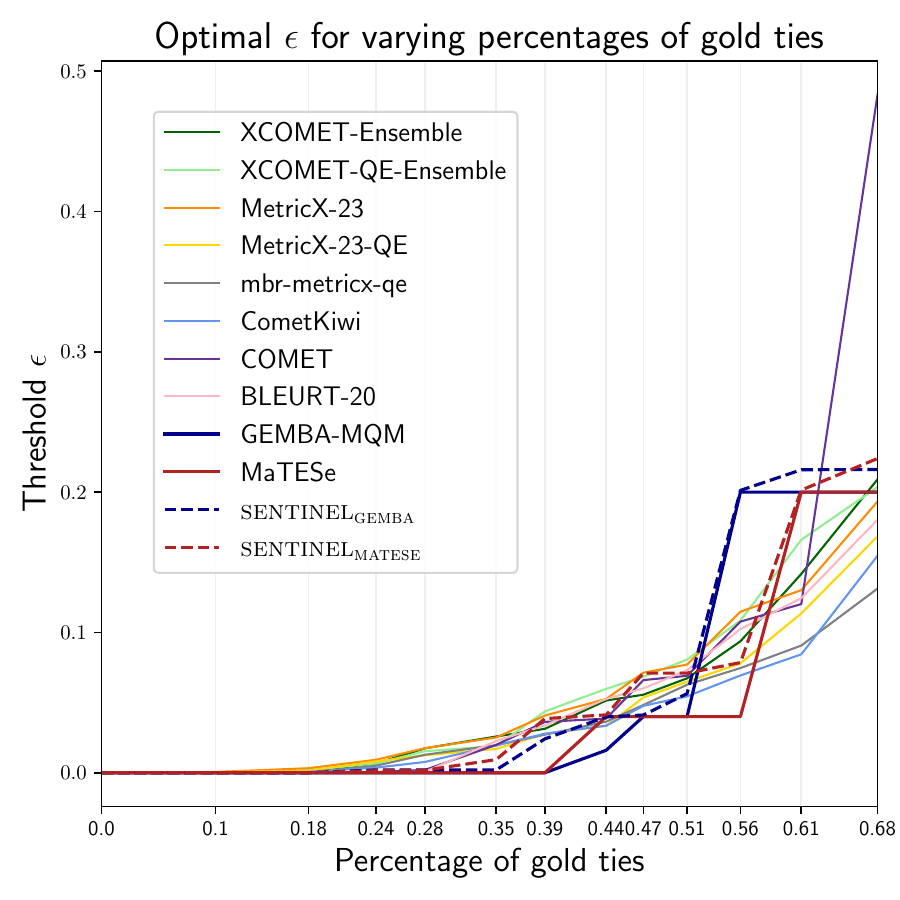}
        \caption{}
        \label{fig:threshold-varying-ties-zhen}
    \end{subfigure}
    }
    \caption{\acceq (a) and optimal $\epsilon$ (b) of the considered metrics for varying percentages of human ties in the test dataset ($0.24$ is the percentage of human ties in the entire dataset, obtained when $p_t$ and $p_n$ are both $0$). $\epsilon$ values have been scaled using min-max scaling. Specifically, for each metric, the minimum $\epsilon$ is the optimal $\epsilon$ at $0\%$ of human ties, and the maximum is the optimal $\epsilon$ at $100\%$. The language direction is \langpair{zh}{en}. For each percentage of human ties, we use $5$ different seeds to sub-sample the test data. Therefore, the shown \acceq and $\epsilon$, for each metric and percentage of ties, are averaged across $5$ different runs.}
    \label{fig:accuracy-and-threshold-varying-ties-zh-en}
\end{figure*}

%% file: figures/accuracy-and-threshold-varying-ties-ende.tex
\begin{figure*}[!ht]
    \centering
    \resizebox{\textwidth}{!}{
    \begin{subfigure}[b]{0.4\linewidth}
        \includegraphics[width=\linewidth]{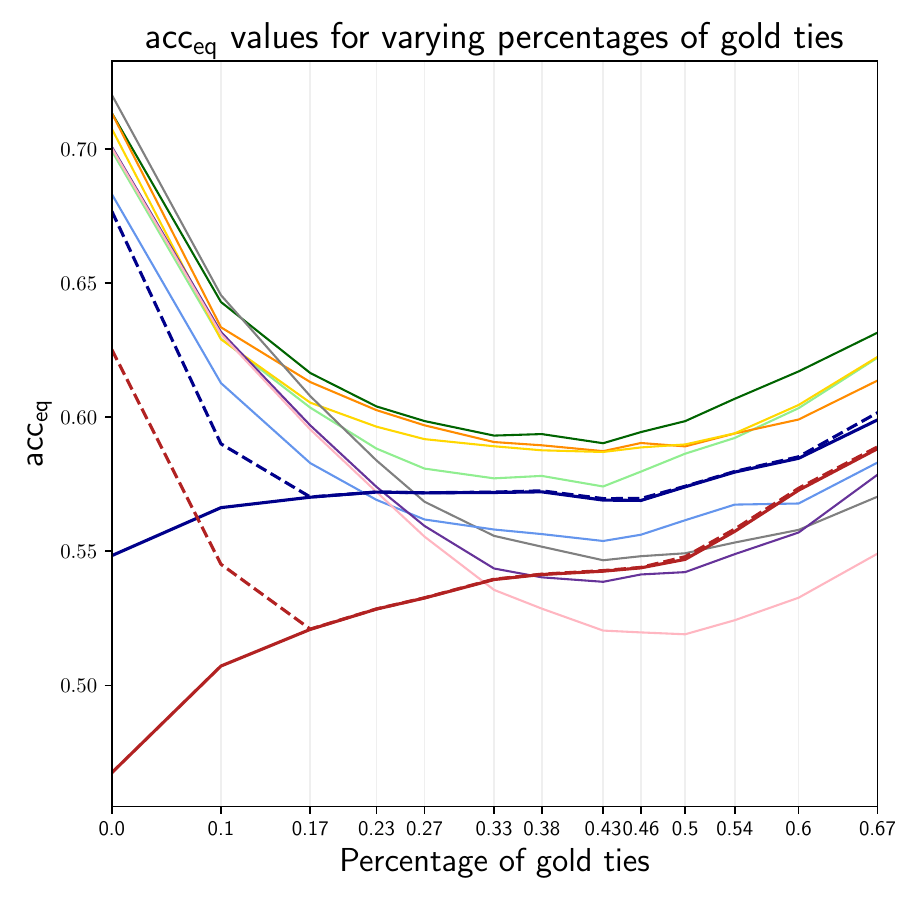}
        \caption{}
        \label{fig:accuracy-varying-ties-ende}
    \end{subfigure}
    \quad
    \begin{subfigure}[b]{0.4\linewidth}
        \includegraphics[width=\linewidth]{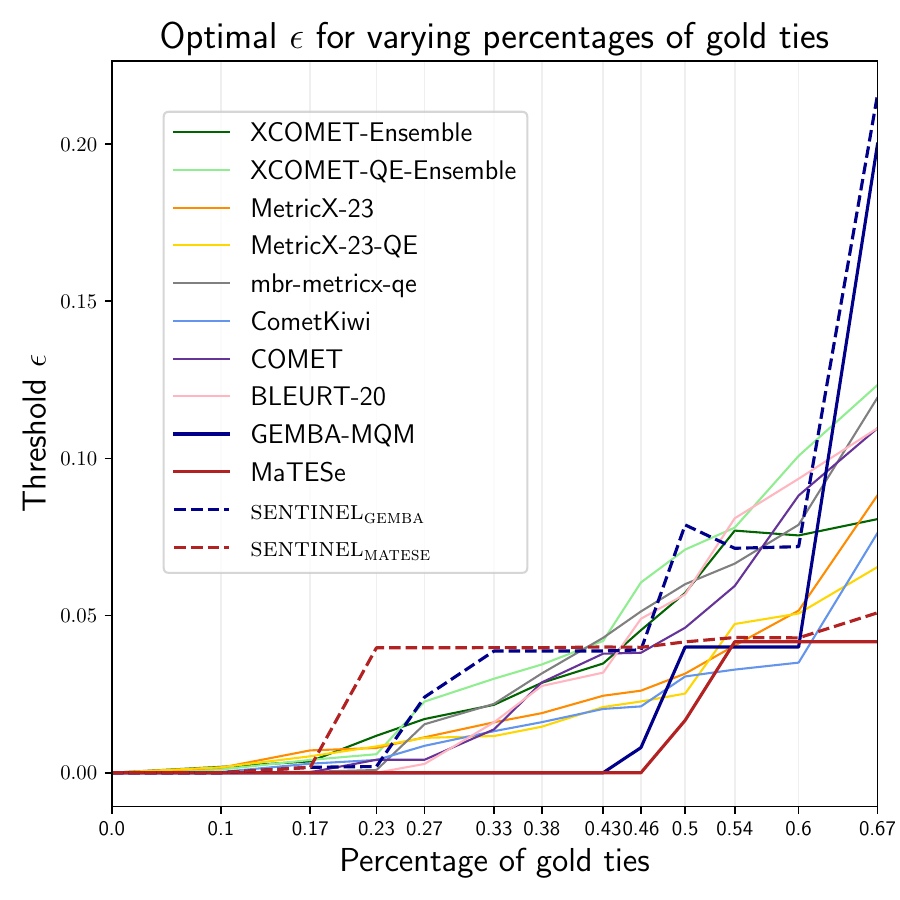}
        \caption{}
        \label{fig:threshold-varying-ties-ende}
    \end{subfigure}
    }
    \caption{\acceq (a) and optimal $\epsilon$ (b) of the considered metrics for varying percentages of human ties in the test dataset ($0.23$ is the percentage of human ties in the entire dataset, obtained when $p_t$ and $p_n$ are both $0$). $\epsilon$ values have been scaled using min-max scaling. Specifically, for each metric, the minimum $\epsilon$ is the optimal $\epsilon$ at $0\%$ of human ties, and the maximum is the optimal $\epsilon$ at $100\%$. The language direction is \langpair{en}{de}. For each percentage of human ties, we use $5$ different seeds to sub-sample the test data. Therefore, the shown \acceq and $\epsilon$, for each metric and percentage of ties, are averaged across $5$ different runs.}
    \label{fig:accuracy-and-threshold-varying-ties-en-de}
\end{figure*}

%% file: figures/accuracy-and-threshold-varying-ties-heen.tex
\begin{figure*}[!ht]
    \centering
    \resizebox{\textwidth}{!}{
    \begin{subfigure}[b]{0.4\linewidth}
        \includegraphics[width=\linewidth]{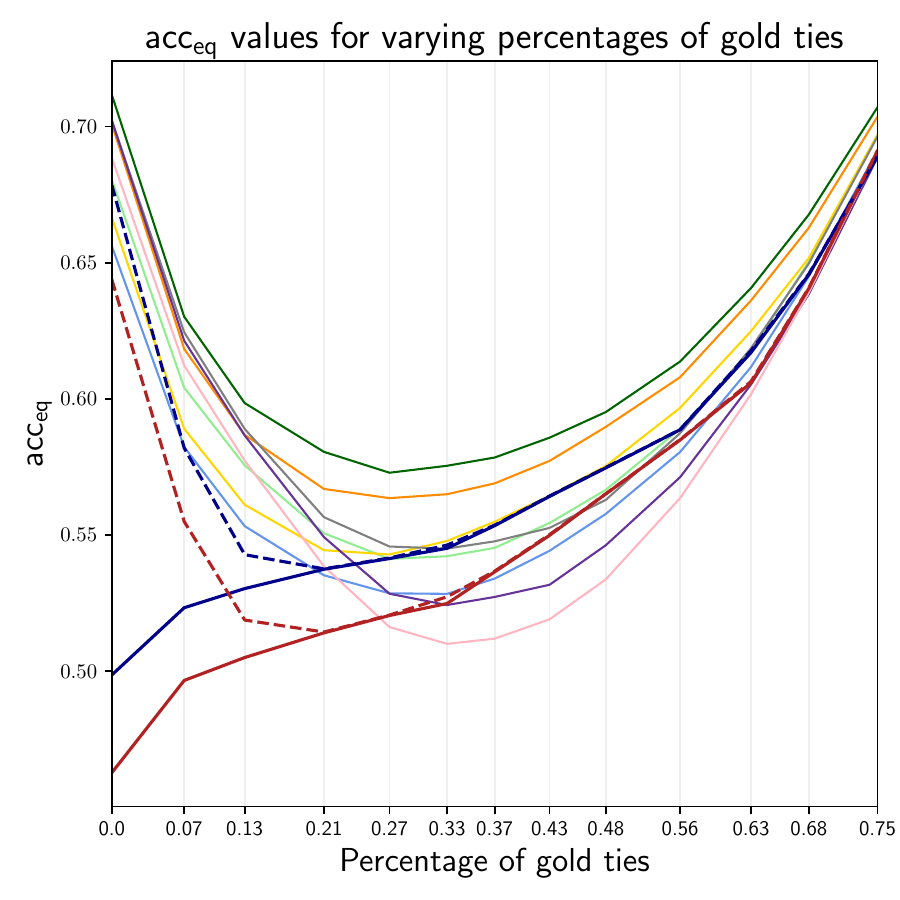}
        \caption{}
        \label{fig:accuracy-varying-ties-heen}
    \end{subfigure}
    \quad
    \begin{subfigure}[b]{0.4\linewidth}
        \includegraphics[width=\linewidth]{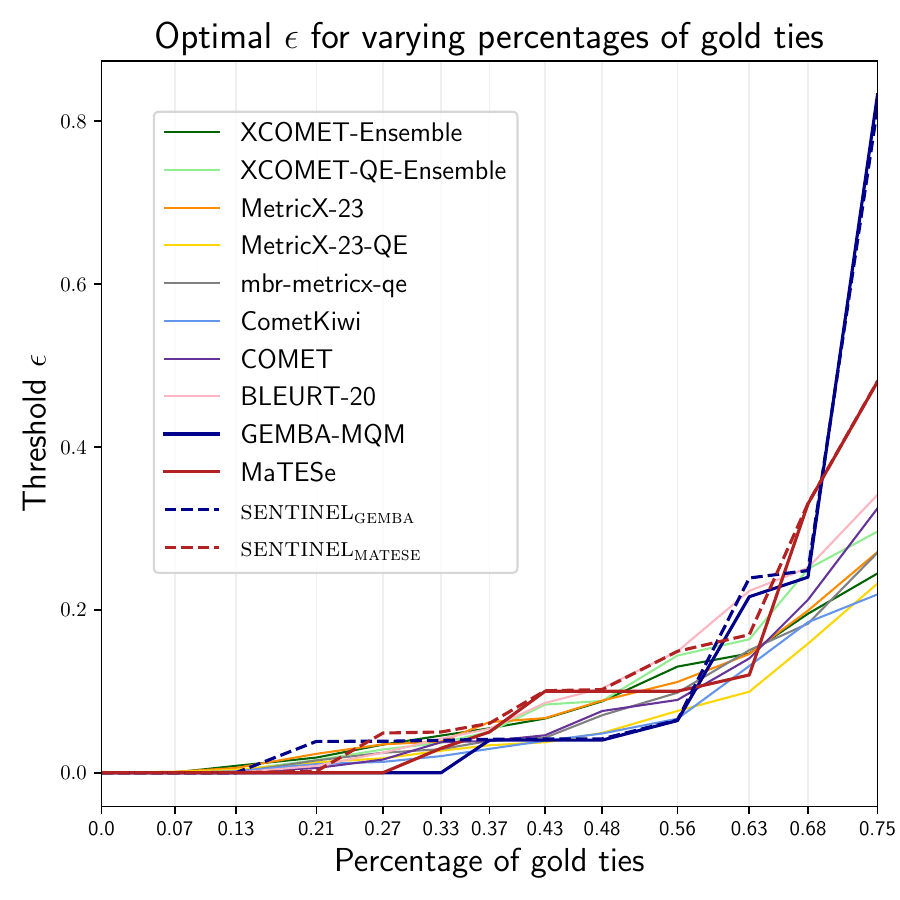}
        \caption{}
        \label{fig:threshold-varying-ties-heen}
    \end{subfigure}
    }
    \caption{\acceq (a) and optimal $\epsilon$ (b) of the considered metrics for varying percentages of human ties in the test dataset ($0.43$ is the percentage of human ties in the entire dataset, obtained when $p_t$ and $p_n$ are both $0$). $\epsilon$ values have been scaled using min-max scaling. Specifically, for each metric, the minimum $\epsilon$ is the optimal $\epsilon$ at $0\%$ of human ties, and the maximum is the optimal $\epsilon$ at $100\%$. The language direction is \langpair{he}{en}. For each percentage of human ties, we use $5$ different seeds to sub-sample the test data. Therefore, the shown \acceq and $\epsilon$, for each metric and percentage of ties, are averaged across $5$ different runs.}
    \label{fig:accuracy-and-threshold-varying-ties-he-en}
\end{figure*}

%% file: figures/dev-accuracy-varying-ties.tex
\begin{figure*}[!ht]
    \centering
    \begin{subfigure}[b]{0.45\linewidth}
        \includegraphics[width=\linewidth]{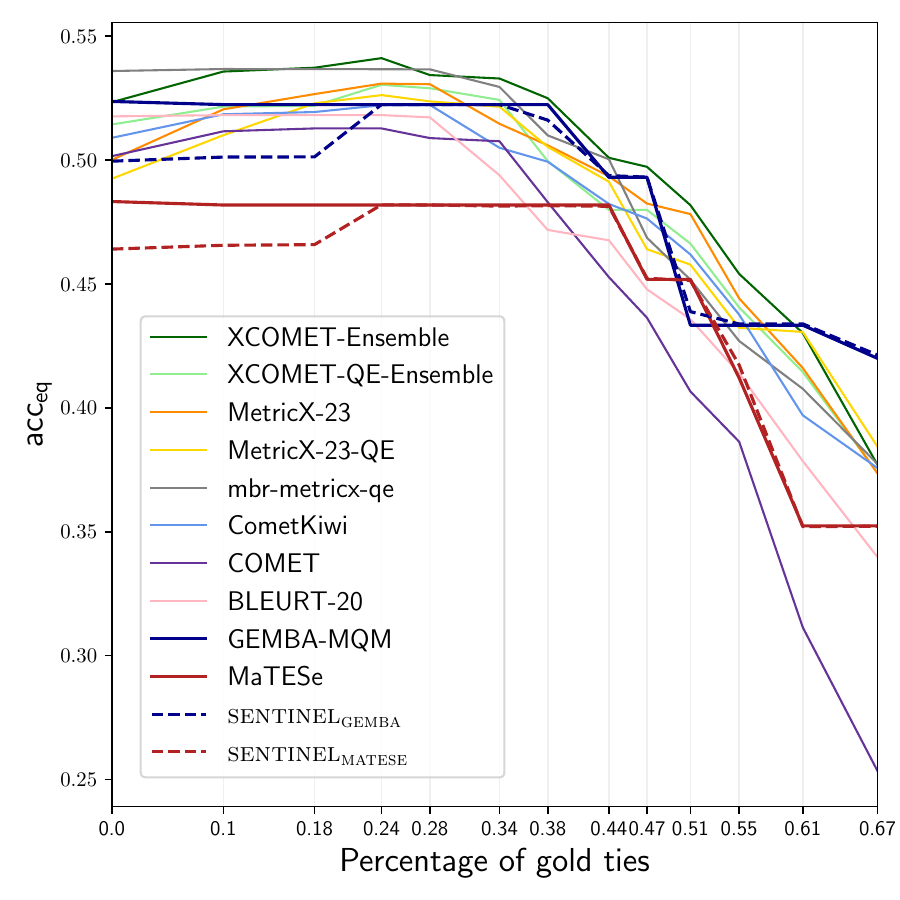}
        \caption{The language pair is \langpair{zh}{en}. The percentage of human ties in the $80\%$ split of the test set is $24\%$. }
        \label{fig:dev-accuracy-varying-ties-zhen}
    \end{subfigure}
    \quad
    \begin{subfigure}[b]{0.45\linewidth}
        \includegraphics[width=\linewidth]{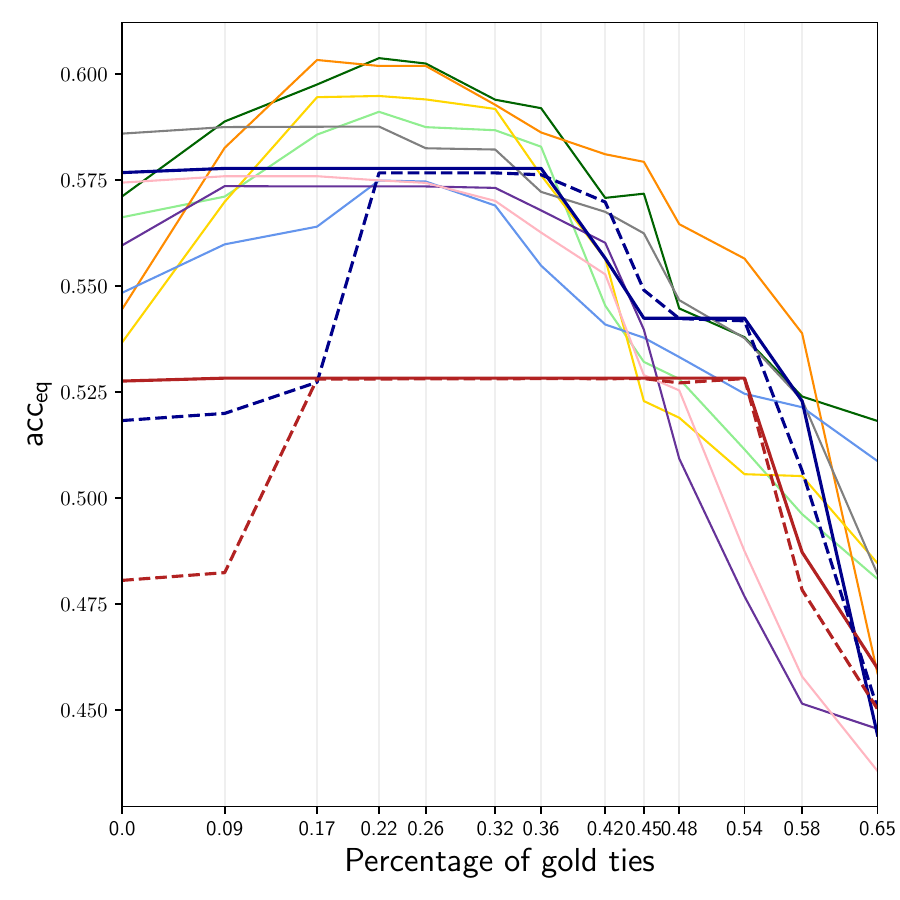}
        \caption{The language pair is \langpair{en}{de}. The percentage of human ties in the $80\%$ split of the test set is $23\%$.}
        \label{fig:dev-accuracy-varying-ties-ende}
    \end{subfigure}
    \begin{subfigure}[b]{0.45\linewidth}
        \includegraphics[width=\linewidth]{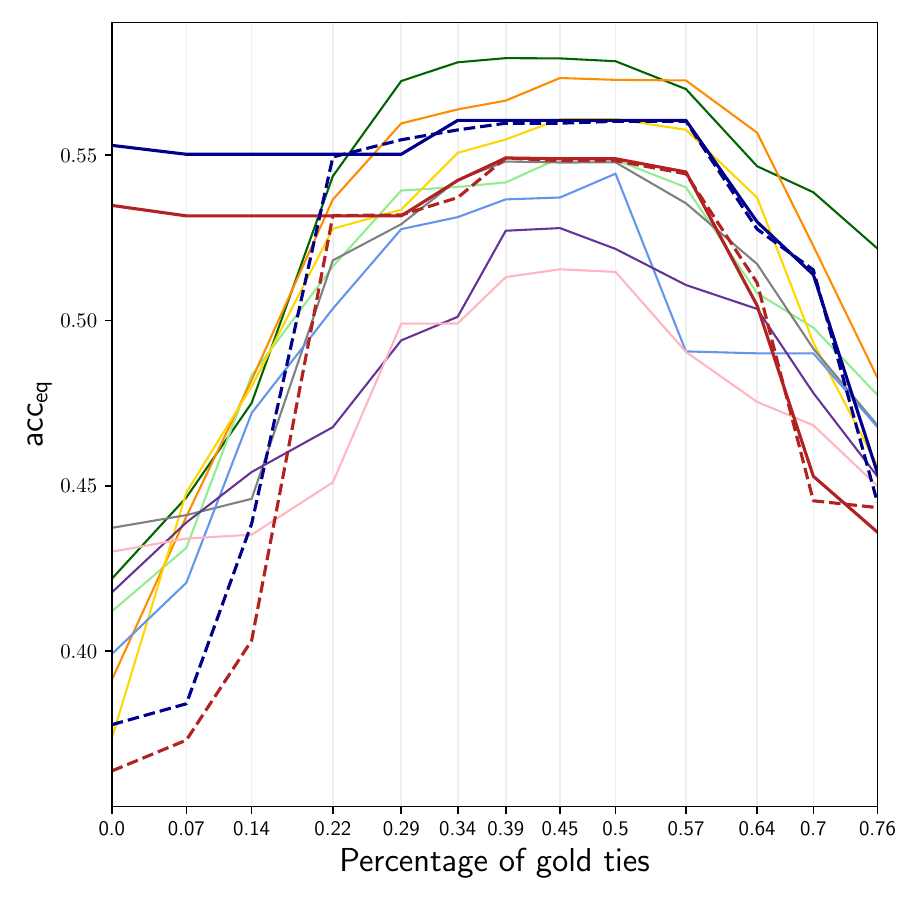}
        \caption{The language pair is \langpair{he}{en}. The percentage of human ties in the $80\%$ split of the test set is $42\%$.}
        \label{fig:dev-accuracy-varying-ties-heen}
    \end{subfigure}
    \caption{\acceq of the considered metrics when tie calibration is conducted on a held-out set, derived as a $20\%$ split of the test set, and repeatedly sub-sampled to modify its percentage of tied scores. The x-axis represents the percentage of ties in the held-out set, while the y-axis represents the \acceq, as computed on the remaining $80\%$ of the test set. For each percentage of human ties, we use $5$ different seeds to sub-sample the held-out set. Therefore, the shown \acceq for each metric and percentage of ties is averaged over $5$ different runs.}
    \label{fig:dev-accuracy-varying-ties}
\end{figure*}